\documentclass{ecai} 

\usepackage{latexsym}
\usepackage{amssymb}
\usepackage{amsmath}
\usepackage{amsthm}
\usepackage{booktabs}
\usepackage{enumitem}
\usepackage{graphicx}
\usepackage{color}
\usepackage{times}
\usepackage{soul}
\usepackage{url}
\usepackage[hidelinks]{hyperref}
\usepackage[utf8]{inputenc}
\usepackage[small]{caption}
\usepackage{graphicx}
\usepackage{amsmath}
\usepackage{amsthm}
\usepackage{booktabs}
\usepackage{algorithm}
\usepackage{algorithmic}
\usepackage{graphicx}
\usepackage{amssymb}
\usepackage{graphicx}
\usepackage{hyperref}
\usepackage{multirow}
\usepackage{verbatim}
\usepackage{hyperref}
\usepackage{subfigure}
\usepackage{booktabs}
\usepackage[table]{xcolor}
\usepackage{multirow}
\usepackage{array}
\usepackage{tabularx}
\usepackage{siunitx}

\newcommand{\BibTeX}{B\kern-.05em{\sc i\kern-.025em b}\kern-.08em\TeX}

\begin{document}


\begin{frontmatter}


\paperid{123} 


\title{DeepDFA: Automata Learning through Neural Probabilistic Relaxations}


\author[A]{\fnms{Elena}~\snm{Umili}\thanks{Corresponding Author. Email: umili@diag.uniroma1.it}}
\author[B]{\fnms{Roberto}~\snm{Capobianco}}

\address[A]{Sapienza University of Rome}
\address[B]{Sony AI}


\begin{abstract}
In this work, we introduce DeepDFA, a novel approach to identifying Deterministic Finite Automata (DFAs) from traces, harnessing a differentiable yet discrete model. Inspired by both the probabilistic relaxation of DFAs and Recurrent Neural Networks (RNNs), our model offers interpretability post-training, alongside reduced complexity and enhanced training efficiency compared to traditional RNNs. Moreover, by leveraging gradient-based optimization, our method surpasses combinatorial approaches in both scalability and noise resilience. Validation experiments conducted on target regular languages of varying size and complexity demonstrate that our approach is accurate, fast, and robust to noise in both the input symbols and the output labels of training data, integrating the strengths of both logical grammar induction and deep learning.
\end{abstract}

\end{frontmatter}


\section{Introduction}
The problem of identifying a deterministic finite state automaton (DFA) from labeled traces is one of the best-studied problems in grammatical inference \cite{grammatical_inference}. The latter sees applications in various areas, including Business Process Management (BPM) \cite{bpm_application}, non-Markovian Reinforcement Learning \cite{apll_1, brafman_ecai_2023}, automatic control \cite{appl_2}, speech recognition \cite{appl_163}, and computational biology \cite{apll_173}. Both passive \cite{sat1} and active \cite{ANGLUIN198787} exact methods have been proposed for DFA identification.
These methods are guaranteed to succeed in theory. However, in practice, they require a notable amount of computation, and they are unable to handle errors in the training dataset, making them of limited applicability, especially to real applications and large DFAs.
Unlike exact approaches, Recurrent Neural Networks (RNN) tolerate errors in the training data, and they have proven highly effective at learning classifiers for sequential data \cite{RNNclass}. 
DFAs and RNNs can both be used for language recognition, which is essentially binary classification over strings. Many works highlight the similarities between RNNs and finite-state machines \cite{RNNpower}. The two differ in the transition representation: RNNs learn a parametrized transition function in a continuous hidden state space, while DFAs have a finite state space and completely transparent transitions. 
Furthermore, designing an RNN requires many choices: deciding the number of layers, the number of features of each hidden layer, and all the activation functions. Each of these decisions can affect the final performance and must be taken carefully. By contrast, exact methods do not require nearly any hyperparameter fine-tuning.
Many approaches have been proposed to extract a DFA from a pre-trained RNN \cite{merrill2022extracting,wang2018,weiss18a_queries}, generally adapting techniques from the grammar induction literature and suggesting ways to discretize or cluster the continuous RNN hidden states in a finite structure. The purpose of these works is not DFA induction but rather to enhance the explainability of black-box sequence classifiers. However, they open the door to DFA induction through neural networks, and join two fields that are classically kept separated. \\
In order to take the benefits from both worlds, grammar induction on one side and recurrent neural networks on the other side, we present DeepDFA: a transparent neural network design specialized in learning DFAs from traces with gradient-based optimization. The model resembles a recurrent neural network, but, differently from RNNs, it is completely transparent after training, as much as a DFA. Furthermore, thanks to its specialized design, it uses fewer weights than the most commonly used recurrent architectures, such as LSTMs and GRUs, and it only has one hyperparameter. This results in faster training and less memory consumption and a significantly reduced hyperparameter search.
At the same time, since it is trained with back-propagation, it is able to learn DFAs in a significantly shorter time than grammar induction methods, even for large automata; and it can tolerate errors in the training labels and the training symbols.
Our method is based on defining a neural network that behaves as a probabilistic finite automaton. We control how much the probabilities are close to categorical one-hot distribution through a temperature value. During training, we smoothly drive the network activations to be close to discrete 0/1 values by changing the temperature. When the gap between the discrete and actual activations is small enough, the network behaves precisely as a DFA. \\
We evaluate our method on the popular Tomita languages benchmark \cite{tomita_lang} and random DFAs of different sizes and different sets of symbols.
Results show that DeepDFA is fast and accurate. It outperforms an exact SAT-based method \cite{Zakirzyanov2019EfficientSB} when the target DFA has more than 20 states, or the training set contains mislabeled examples, and it reaches competitive results in the other cases. We also compared DeepDFA to DFA extraction from a pre-trained RNN \cite{weiss18a_queries}, finding it reaches better accuracy and predicts DFA of size closer to the target DFA size.
The remainder of this paper is organized as follows: in section 2 we report related works; in section 3 we give some preliminaries on Deterministic and Probabilistic Finite Automata and Recurrent Neural Networks; in section 4 we formulate our problem and illustrate in detail the framework used to solve it; we report the experiments evaluating our approach in section 5; and finally we conclude and discuss directions for future work in section 6.
\section{Related works} \label{sec:related_works}
\paragraph{Combinatorial methods for grammar induction}
Many approaches have been proposed to identify a target DFA from a set of positive and negative traces.
The L$^*$ algorithm \cite{ANGLUIN198787}, is an exact active learning algorithm to learn a DFA from an oracle through membership and equivalence queries.
Another approach is to apply the evidence-driven state-merging algorithm \cite{state_merging1,state_merging2}, which is a greedy algorithm that is not guaranteed to converge to the global optimum.
A more modern approach is to leverage highly-optimized SAT solvers \cite{sat1}. This approach is guaranteed to find the minimal DFA consistent with all the training examples, but suffers from scalability problems. In this regard, several symmetry-breaking predicates have been proposed for the SAT encoding to reduce the search space \cite{Zakirzyanov2017FindingAM,Zakirzyanov2019EfficientSB}. 
\paragraph{DFA extraction from recurrent neural networks}
Prior works extract a DFA from a pre-trained RNN, to explain the network behavior.
Weiss et al. \cite{weiss18a_queries} adapt the L$^*$ algorithm \cite{ANGLUIN198787} to work with an RNN oracle. 
Other work uses k-means clustering on the RNN hidden states to extract a graph of states \cite{wang2018}. Merrill et al. extract a DFA from an RNN by first building a prefix tree and then merging states with close state representation in the RNN hidden space \cite{merrill2022extracting}.
These approaches train an RNN with the labeled strings and then extract an equivalent DFA from the trained model. 
Our approach differs from these since we directly train an RNN equivalent to a Probabilistic Finite Automaton (PFA), and we force the probabilities to become close to one-hot categorical distributions \textit{during training}. In this way, our model \textit{becomes a DFA}. In other words, there is no difference between the trained neural model and the automaton, and there is no risk that the abstraction does not represent the network, as for previous works. 
For example, Wang et al. \cite{wang2018} cluster the RNN states, so the automaton depends on the clustering algorithm performance. Performances from \cite{merrill2022extracting} instead rely on a similarity threshold and, as the paper shows, also on the number of epochs the RNN is trained after convergence. Our work in this sense is more similar to \cite{weiss18a_queries}, since it computes the abstraction \textit{automatically}. The difference is in how the abstraction is computed: we use gradient descent optimization while \cite{weiss18a_queries} starts with a hand-crafted discretization that is automatically refined during automata learning with L*.
\paragraph{Learning crispy logical models through gradient descent}
Classically, the induction of logic models, including DFAs, is not approached with gradient-descent optimization methods (as deep learning methods) since their finite and `crispy' nature prevents the gradient computation.
However, recent works in neurosymbolic AI \cite{thirdvawe} propose techniques to reduce the gap between the induction of crispy models and that of continuous ones.
Walke et al. \cite{walke_ltl} proposes a recurrent neural model with specialized filters to learn Linear Temporal Logic formulas over finite traces (LTLf) from labeled traces. Collery et al. \cite{Collery_rule_induction} discovers logical rules describing patterns in sequential data using a differentiable model. Aichernig et al. \cite{rew3} learn the DFA by constraining the outputs of a simple RNN to be close to one-hot vectors through a regularization term in the loss. Differently from \cite{rew3}, we define a new recurrent model specialized in learning PFAs, which we discretize at training time with temperature annealing and train with the classical cross-entropy between the predicted and the ground-truth label.
Grachev et al. \cite{petr2017} proposes a neural network model similar to ours to learn DFA from traces. However, this model does not use activation functions with discrete outputs, and it is affected by the vanishing gradient problem for automata larger than 5 states, while our method can effectively learn target automata with up to 30 states.
\section{Background}
\begin{figure*}[t!]
    \centering
    \subfigure[]{
    \includegraphics[width=0.41\textwidth]{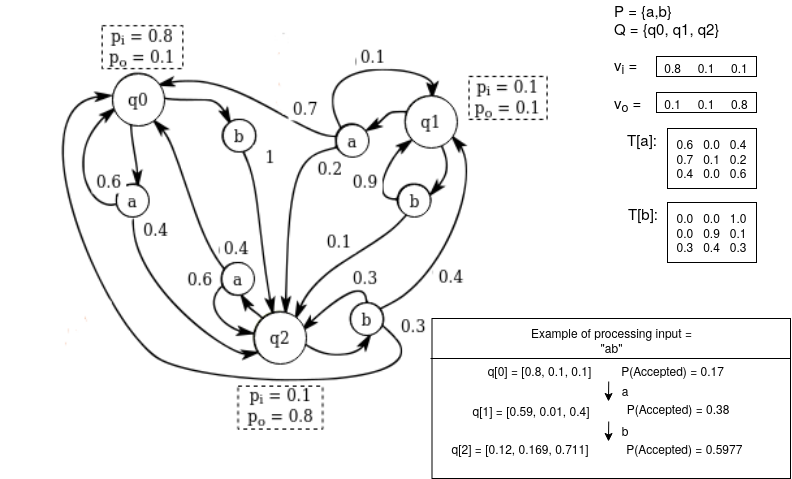}}
    \subfigure[]{
    \includegraphics[width=0.27\textwidth]{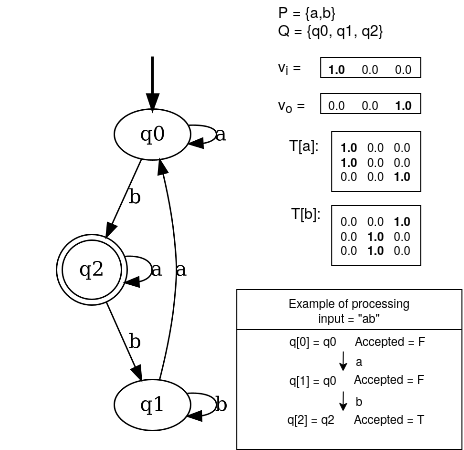}}
    \subfigure[]{
    \includegraphics[width=0.29\textwidth]{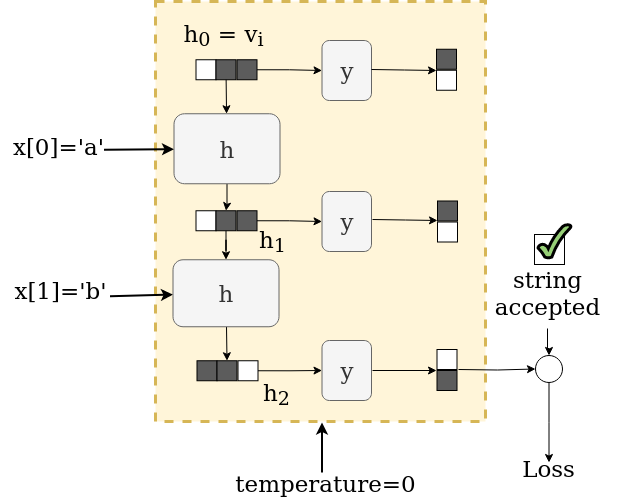}}
    \caption{a) An example of PFA with three states and two symbols: graph describing the PFA, equivalent representation in matrix form, and produced states and acceptance probabilities while processing the string "ab". b) An example of DFA: graph describing the PFA, equivalent representation in matrix form, and produced states and acceptance probabilities while processing the string "ab". In particular, the DFA in (b) is obtained by the PFA in (a) approximating the matrix representation to the closest one-hot vectors. c) DeepDFA processing the string "ab".}
    \label{fig:PFAvsDFA}
\end{figure*}
\subsection{Deterministic Finite Automata}
A Deterministic Finite Automaton (DFA) $A$ is a tuple $(P, Q, \delta, q_0,  F)$, where $P$ is a set of propositional symbols called the alphabet, $Q$ is a finite set of states, $q_0 \in Q$ is the initial state, $\delta: Q \times P \to Q $ is the transition function, and $F \subseteq Q$ is the set of final states.
Let $P^*$ be the set of all finite strings over $P$, and $\epsilon$ the empty string.
The transition functions over strings $\delta^*:Q \times P^* \rightarrow Q$ is defined recursively as
\begin{equation}
\begin{array}{l}
    \delta^*(q,\epsilon) = q \\
    \delta^*(q, ax) = \delta^*(\delta(q,a) , x)
\end{array}
\end{equation}
Where $a \in P$ is a symbol and $x \in P^*$ is a string, and $ax$ is the concatenation of $a$ and $x$.
$A$ accepts the string $x$ (i.e., $x$ is in the language of $A$, $L(A)$) if and only if $\delta^*(q_0, x) \in F$.
Let $x=x[1]x[2]...x[l]$ be the input string, where $x[t]$ is the $t$-th character in the string, we denote as $q =q[0]q[1]...q[l]$ the sequence of states visited by the automaton while processing the string, namely $q[0] =q_0$ and $q[t]=\delta(q[t-1],x[t])$ for all $t > 0$.

\subsection{Probabilistic Finite Automata} \label{sec:PFA}
A Probabilistic Finite Automaton (PFA) $A_p$ is a tuple $(P, Q, i_{p}, \delta_{p}, f_p)$, where $P$ is the alphabet, $Q$ is the finite set of states, $i_{p}: Q \rightarrow [0,1]$ is the probability for a state to be an initial state, $\delta_{p}: Q \times P \times Q \rightarrow [0,1]$ is the transition probability function, and $f_p:Q \rightarrow [0,1]$ is the probability of a state to be final. We have therefore $\sum\limits_{q' \in Q}\delta_{p}(q,p,q') = 1$, and $\sum\limits_{q \in Q}i(q)=1$ $\forall q \in Q$, $\forall a \in P$.

We can represent the PFA in matrix form as a \textit{transition matrix} $T$, an \textit{input vector} $v_i$ and an \textit{output vector} $v_o$.
Matrix $T \in \mathbb{R}^{|P|\times|Q| \times |Q|}$ contains at index $(p, q, q')$ the value of $\delta_{p}(q, p, q')$. We denote as $T[p] \in \mathbb{R}^{|Q| \times |Q|}$ the 2D transition matrix for symbol $p$. 

The input vector $v_i \in \mathbb{R}^{1 \times |Q|}$ contains at index $k$ the probability of state $q_k$ to be an initial state, while the output vector $v_o \in \mathbb{R}^{|Q| \times 1}$ has in position $k$ the probability of state $q_k$ to be accepting. This matrix representation is shown in Figure \ref{fig:PFAvsDFA}(b).

Given a string $x=x[0]x[1]...x[l-1]$, we denote as $q_{p,0}, q_{p,1}...q_{p,l}$ the sequence of probabilities to visit a certain state, where $q_{p,t} \in \mathbb{R}^{1 \times |Q|}$ is a row vector containing at position $k$ the probability to stay in state $k$ at time $t$. 
\begin{equation}
\begin{array}{l}
    q_{p,0} = v_i \\
    q_{p,t} = q_{p,t-1} T[x[t]]\quad\quad \forall t > 0
\end{array}
\end{equation}
The probability of being in a final state at time $t$ is the inner product $q_{p,t} v_o$.

Therefore the probability of a string to be accepted is the probability to be in a final state in the last computed state $q_{p,l}$, and it is calculated as follows
\begin{equation}
    v_i T[x[0]] T[x[1]]..  T[x[l-1]] v_o\label{acceptance_formula}
\end{equation}
Figure \ref{fig:PFAvsDFA} shows a comparison between a DFA and a PFA.

\subsection{Recurrent Neural Networks}
A Recurrent Neural Network (RNN) is a parameterized function $h_t = f(h_{t-1}, x_{t}; \theta_h)$, having trainable parameters $\theta_h$, that takes as input a state-vector at time $t-1$, $h_{t-1}$, and the input vector at time $t$, $x_{t}$, and returns the current state-vector at time $t$, $h_{t}$. An RNN can be applied to a sequence $x[0] , ..., x[n]$ by recursive application of the function $h$ to the vectors $x[i]$.  
An example of RNNs are the Elman and Jordan networks \cite{simple_recurrent_networks}, also known as Simple Recurrent Networks.
\begin{equation}
\begin{array}{l}
    h_t = \sigma_h(W_hx_t + U_hh_{t-1} + b_h)\\
    y_t = \sigma_y(W_yh_t + b_y)\\
\end{array}
\end{equation}
Where $x$ is the network input, $h$ is the hidden state, $y$ is the network output, $W_h, U_h, W_y$ and $b_h$ are the network parameters, and $\sigma_h$ and $\sigma_y$ are activation functions.

RNNs are employed in a variety of tasks on sequential data, including sequence classification. A binary RNN-acceptor, is an RNN where $y$ classifies the RNN’s state vectors as accepting or rejecting. An RNN-acceptor is conceptually equal to a DFA, except that it processes a sequence by applying continuous state-transitions and output evalutions, that are usually hard to inspect and explain. In our work we propose an alternative RNN model tailored on DFAs that makes the model interpretable by design.


\section{DeepDFA}
We consider the problem of inferring a DFA from a training set of labeled strings $D= \{(x_1, \bar{y_1}), (x_2,\bar{y_2}), ..., (x_n, \bar{y_n})\}$, where $x_i$ is a string of length $l$ of symbols $x[0], x[1], ...,x[l-1]$, in the automaton alphabet $P$, $x_i[t] \in P\text{ with } 0 \leq t < l$, and $ \bar{y_i}\in \{0, 1\}$ is the associated label, denoting whether the string is accepted or not by the target automaton. 

To infer the automaton, we define a recurrent neural network model that mimics the behavior of a PFA in a state space $\hat{Q}_{max}$ and action space $P$, where $P$ is the target automaton alphabet while $\hat{Q}_{max}$ is our hypothesis on the state space, which will generally differ from the true $Q$. For this reason, the size of $\hat{Q}_{max}$ is a hyperparameter of our model.

We cannot use gradient-based optimization to learn the DFA directly because of its non-differentiable transitions and output vector.
The intuition behind our work is that PFAs are closely related to recurrent neural networks, since they calculate the next state and output using multiplications between continuous vectors and matrices, in the same way RNNs do. However, DFAs can also be represented in matrix form, with the difference that their matrix representation is composed of only one-hot row vectors. As example we show in Figure \ref{fig:PFAvsDFA} a simple PFA (subfigure (a)) and the DFA obtained by approximating all the PFA matrix representation row vectors to the closest one-hot (subfigure (b)).

Following this idea, we define DeepDFA as a parametric PFA in which we can drive the representation to be close to one-hot during training.
We obtain this effect using an activation function that smoothly approximates discrete output. Many works use this technique \cite{gumbel_softmax,walke_ltl,marine_rule_induction}, especially in neurosymbolic AI \cite{nesy21}, to learn symbolic logic structures by using differentiable models such as neural networks.
In particular, we use a modified version of the classical softmax and sigmoid activation functions, which we call $softmax\_with\_temp$ and $sigmoid\_with\_temp$.
Given a function $f(x)$ we define $f\_with\_temp(x, \tau) = f(x/\tau)$, with $\tau$ being a positive temperature value.

Our RNN comprises two trainable modules: a transition function and an output function.
The transition function $h_t(h_{t-1},x[t-1]; \theta_h)$ has parameters $\theta_h$, takes as input the probabilities over the previous state, $h_{t-1}\in [0,1]^{|\hat{Q}|_{max}}$, and the previous symbol, $x[t-1] \in P$, and predicts the current state $h_{t}$.
The output function $y(h_t; \theta_y)$ implements the classification module: given the current state estimation, $h_t$, predicts the probability of the current state to be an accepting state using its parameters $\theta_y$.
In particular, fixed a temperature value $\tau$
\begin{equation} \label{eq:transition_rnn}
\begin{array}{l}
    h_0 = v_i = [1,0,...,0] \\
    h_{t} = h_{t-1} T[x[t-1]]\\
    y_t = h_t v_o \\
    T = softmax\_with\_temp(\theta_h, \tau) \\
    v_o = sigmoid\_with\_temp(\theta_y, \tau) \\
\end{array}
\end{equation}
Where $T$ is the PFA transition matrix, $T[x[t]]$ is the transition matrix for symbol $x[t]$, and $v_o$ is the output vector, as defined in Section \ref{sec:PFA}. They are obtained by applying discrete activation functions on parameters $\theta_h \in \mathbb{R}^{|P|\times |\hat{Q}_{max}|\times|\hat{Q}_{max}|}$ and $\theta_y \in \mathbb{R}^{{|\hat{Q}_{max}| \times 1}}$.
In particular, the softmax activation applied to the third dimension of the matrix $\theta_h$ ensures $\sum_{q'}\delta_p(q,a,q') = 1$, and the sigmoid activation ensures values of $v_o$ stay in $[0,1]$.
 
In the forward pass, we calculate the probability of a string $x$ in the dataset to be accepted by the RNN by applying Equation \ref{eq:transition_rnn} recursively to the input sequence of symbols $ x[0], x[1], ..., x[l-1]$.
The final output $y_{l}$ is compared with the ground truth label $\bar{y}$.
 We update the model weights with back-propagation by minimizing the binary cross-entropy between model predictions and the ground truth labels.

\paragraph{Temperature Annealing}
Cold temperatures force the PFA to behave as a DFA, since the activation values become closer to boolean values as the temperature decreases. When the temperature is low enough, the model transforms into a DFA.
Let us notice that using the classical activation functions and discretizing the model at the end is not guaranteed to work because the discretized model differs from the trained one, and the two can have different performances.
%
However, using a cold temperature from the start of training can prevent the system from learning the model properly.
For this reason, we initialize $\tau$ to 1. In this way, the system starts the training using the normal softmax and activation functions. After that, we anneal the temperature to 0 by multiplying it by a positive constant  $\lambda < 1$ at each epoch.
In Figure \ref{fig:PFAvsDFA}(c), we show the network behavior in case of a perfectly discretized activation, namely when the temperature is 0. In theory, this ideal behavior is obtained only with zero temperature, but in practice, we observe the continuous and discretized models start having the same accuracy already at temperatures around 0.8.
\paragraph{Model Minimization}
Once the training is concluded, we read the DFA from the activations evaluated with the minimum temperature, and we use the Hopcroft algorithm \cite{minimization_alg} to minimize the number of states. 
We emphasize that automata minimization is a well-known problem for which many algorithms are available, as opposed to neural network compression, also known as knowledge distillation \cite{distillation}, which is still an open research problem. This represents another feature in which DeepDFA can take `the best of both worlds.'
We observe that, even if the state space size $|\hat{Q}_{max}|$ is large, the number of states after minimization $|\hat{Q}|$ tends to be close to the target DFA number of states, which suggests the model is robust to overfitting.
\paragraph{Extension to Probabilistic Symbols} \label{sec:extension}
DeepDFA  allows us to represent uncertainty over both the transition and the output function, because it is based on representing the model as a probabilistic machine.
However, let us notice the current symbol $x[i]$ is used to index the transition matrix in Equations \ref{eq:transition_rnn} and, as a consequence, \textit{symbols must be integers}. This contrasts with symbol grounding techniques based on neural networks, which usually predict a \textit{probabilistic belief} on symbol truth values.
For this reason, we extend the framework to be fully probabilistic, and consider probability values over symbols in the calculations.
Given a sequence of inputs $x_1,x_2,...,x_l$, where $x_i$ is a probability vector over $|P|$ classes, we define the next state and output of the network as follows
\begin{equation} \label{eq:deepDFA_uncertain_sym}
\begin{array}{l}
    h_0 = v_i = [1,0,...,0] \\
    h_{t}= \sum\limits_{p=0}^{p=|P|} x_{t-1,p}(h_{t-1} T[p] )\\
    y_t = h_t v_o \\
    T = softmax\_with\_temp(\theta_h, \tau) \\
    v_o = sigmoid\_with\_temp(\theta_y, \tau) \\
\end{array}
\end{equation}
where we denote as $x_{t,p}$ the probability that $x_t$ is symbol $p \in P$.
\begin{table*}[t!]
\caption{Comparison between\textbf{ DeepDFA}, \textbf{L* extraction}, \textbf{DFA-inductor} and \textbf{DFA-generator} on the \textbf{Tomita} Languages. We report test accuracy, mean number of states $|\hat{Q}|$, and the number of parameters used $\#W$.}
\label{tbl:comparison_tomita}
\centering
\begin{tabular}{lcccccccccccc}
\toprule
Lang & \multicolumn{3}{c}{DeepDFA} & \multicolumn{3}{c}{L* extraction} & \multicolumn{2}{c}{DFA-inductor} & \multicolumn{3}{c}{DFA-generator} \\
\cmidrule(lr){2-4} \cmidrule(lr){5-7} \cmidrule(lr){8-9} \cmidrule(lr){10-12}
 & Accuracy & $|\hat{Q}|$ & $\#W$ & Accuracy & $|\hat{Q}|$ & $\#W$ & Accuracy & $|\hat{Q}|$  & Accuracy & $|\hat{Q}|$ & $\#W$\\
\midrule
T1 ($|Q|=2$) & \textbf{100 $\pm$ 0} & \textbf{2 $\pm$ 0} & 220 & \textbf{100 $\pm$ 0} & \textbf{2 $\pm$ 0} & 3030 & \textbf{100 $\pm$ 0} & \textbf{2}  & 100 $\pm$ 0  & 1.8 $\pm$ 1.1 & \textbf{210}\\
T2 ($|Q|=3$) & \textbf{100 $\pm$ 0} & \textbf{3 $\pm$ 0} & 220 & \textbf{100 $\pm$ 0} & \textbf{3 $\pm$ 0} & 3030 & \textbf{100 $\pm$ 0} & \textbf{3}  & 100 $\pm$ 0 & 1 $\pm$ 0 & \textbf{210}\\
T3 ($|Q|=5$) & \textbf{100 $\pm$ 0} & \textbf{5 $\pm$ 0} & 220 & \textbf{100 $\pm$ 0} & \textbf{5 $\pm$ 0} & 3030 & \textbf{100 $\pm$ 0} & \textbf{5}  & 91 $\pm$ 15 & 5 $\pm$ 1.6 & \textbf{210}\\
T4 ($|Q|=4$) & \textbf{100 $\pm$ 0} & \textbf{4 $\pm$ 0} & 220 & \textbf{100 $\pm$ 0} &\textbf{ 4 $\pm$ 0} & 3030 & \textbf{100 $\pm$ 0} & \textbf{4}  & 85 $\pm$ 33 & 4.8 $\pm$ 0.4 & \textbf{210}\\
T5 ($|Q|=4$) & \textbf{100 $\pm$ 0} & \textbf{4 $\pm$ 0} & 220 & \textbf{100 $\pm$ 0} & \textbf{4 $\pm$ 0} & 62000 & \textbf{100 $\pm$ 0} &\textbf{ 4}  & 100 $\pm$ 0 & 2.6 $\pm$ 0.9 & \textbf{210}\\
T6 ($|Q|=3$) & \textbf{100 $\pm$ 0} & \textbf{3 $\pm$ 0} & 220 & 91.1 $\pm$ 19.8 & 170.6 $\pm$ 374.7 & 41400 & \textbf{100 $\pm$ 0} & \textbf{3}  & \textbf{100 $\pm$ 0} & \textbf{3 $\pm$ 0} & \textbf{210}\\
T7 ($|Q|=5$) & \textbf{100 $\pm$ 0} & \textbf{5 $\pm$ 0} & 220 & \textbf{100 $\pm$ 0} & \textbf{5 $\pm$ 0} & 3030 & \textbf{100 $\pm$ 0} & \textbf{5}  & 100 $\pm$ 0 & 4.4 $\pm$ 0.9 & \textbf{210}\\
\bottomrule
\end{tabular}
\end{table*}

\begin{table*}[t!]
\caption{Comparison between \textbf{DeepDFA}, \textbf{L* extraction},\textbf{DFA-inductor} and \textbf{DFA-generator} on the \textbf{noisy dataset} constructed for the \textbf{Tomita} Languages. We report test accuracy, mean number of states $|\hat{Q}|$, and the number of parameters used $\#W$.}
\label{tbl:comparison_tomita_noisy}
\centering
\begin{tabular}{lcccccccccccc}
\toprule
Lang & \multicolumn{3}{c}{DeepDFA} & \multicolumn{3}{c}{L* extraction} & \multicolumn{2}{c}{DFA-inductor} & \multicolumn{3}{c}{DFA-generator} \\
\cmidrule(lr){2-4} \cmidrule(lr){5-7} \cmidrule(lr){8-9}\cmidrule(lr){10-12}
 & Accuracy & $|\hat{Q}|$ & $\#W$ & Accuracy & $|\hat{Q}|$ & $\#W$ & Accuracy & $|\hat{Q}|$ &  Accuracy & $|\hat{Q}|$ & $\#W$\\
\midrule
T1 ($|Q|=2$) & \textbf{100 $\pm$ 0} & \textbf{2 $\pm$ 0} & 1860 & \textbf{100 $\pm$ 0} & \textbf{2 $\pm$ 0} & 3030 & 95 & 12 & 100 $\pm$ 0 & 2.2 $\pm$ 1.1 & \textbf{210}\\
T2 ($|Q|=3$) & \textbf{100 $\pm$ 0} & \textbf{3 $\pm$ 0} & 220 & \textbf{100 $\pm$ 0} & \textbf{3 $\pm$ 0} & 4020 & 94 & 10  & 100 $\pm$ 0 & 1 $\pm$ 0 & \textbf{210}\\
T3 ($|Q|=5$) & \textbf{100 $\pm$ 0} & \textbf{5 $\pm$ 0} & 220 & 100 $\pm$ 0 & 7.4 $\pm$ 3.3 & 4020 & o.o.m.& -- & 100 $\pm$ 0 & 4.8 $\pm$ 0.8 & \textbf{210}\\
T4 ($|Q|=4$) & \textbf{100 $\pm$ 0} & \textbf{4 $\pm$ 0} & \textbf{220} & \textbf{100 $\pm$ 0} & \textbf{4 $\pm$ 0} & 4020& o.o.m.& -- & 90 $\pm$ 22.3 & 5.8 $\pm$ 1.3 & 1830 \\
T5 ($|Q|=4$) & \textbf{100 $\pm$ 0} & \textbf{4 $\pm$ 0} & 220 & \textbf{100 $\pm$ 0} & \textbf{4 $\pm$ 0} & 62000& o.o.m.& --  & 100 $\pm$ 0 & 2.6 $\pm$ 0.8 & \textbf{210}\\
T6 ($|Q|=3$) & \textbf{100 $\pm$ 0} & \textbf{3 $\pm$ 0} & 220 & 99.9 $\pm$ 0.1 & 107 $\pm$ 170.6 & 41400& o.o.m.& --  & \textbf{100 $\pm$ 0} & \textbf{3 $\pm$ 0} & \textbf{210}\\
T7 ($|Q|=5$) & \textbf{100 $\pm$ 0} & \textbf{5 $\pm$ 0} & 220 & \textbf{100 $\pm$ 0} & \textbf{5 $\pm$ 0} & 31100& o.o.m.& --  & 100 $\pm$ 0 & 4.4 $\pm$ 1.1 & \textbf{210}\\
\bottomrule
\end{tabular}
\end{table*}

\section{Experimental Evaluation}
In this section, we report the main results supporting our method. We provide our code at \href{https://github.com/whitemech/DeepDFA}{https://github.com/whitemech/DeepDFA}.
\paragraph{Target DFAs}
We test our approach on two different sets of DFAs. The first experiment is on the Tomita languages \cite{tomita_lang}, which are a standard benchmark for grammar induction and DFA extraction from RNNs \cite{weiss18a_queries,wang2018}.
The benchmark comprises seven formal languages of increasing difficulty defined on the binary alphabet $P=\{a,b\}$.
Despite Tomita languages being a popular benchmark, they are represented by small DFAs with state size smaller than six. In order to test our approach on bigger DFAs, we conduct a second experiment on randomly generated DFAs with state size $|Q|$ between 10 and 30, and alphabet size $|P|$ between 2 and 3. For each setting, we generate 5 random DFAs as described in \cite{Zakirzyanov2017FindingAM}. $|Q|$ states are generated and enumerated between 1 and $|Q|$, we set 1 as the initial state, and each state is equiprobable to be accepting. We connect each state $i$ with a random state in $[i+1, Q]$ with a random-labeled transition. In this way, we partially build an automaton where all states are reachable from the initial state. Finally, we complete the DFA with random transitions.
    
\begin{table*}[t!]
\caption{Comparison between \textbf{DeepDFA}, \textbf{L* extraction}, \textbf{DFA inductor} and \textbf{DFA-generator} on \textbf{random DFAs}. We keep the best 5 experiments over 10 for for the stochastic methods.}
\label{tbl:comparison_random_dfa}
\centering
    \setlength{\tabcolsep}{2pt}

\begin{tabular}{lcccccccccccccc}
\toprule
\multirow{2}{*}{$|Q|$} & \multirow{2}{*}{$|P|$} & \multicolumn{3}{c}{DeepDFA} & \multicolumn{3}{c}{L* extraction} & \multicolumn{3}{c}{DFA-inductor} & \multicolumn{3}{c}{DFA-generator} \\
\cmidrule(lr){3-5} \cmidrule(lr){6-8} \cmidrule(lr){9-11} \cmidrule(lr){12-14}
& & Test Acc. & $|\hat{Q}|$ & Exec. Time & Test Acc. & $|\hat{Q}|$ & Exec. Time & Test Acc. & $|\hat{Q}|$ & Exec. Time & Test Acc. & $|\hat{Q}|$ & Exec. Time \\
\midrule
10 & 2 & \textbf{100 $\pm$ 0} & \textbf{8.0 $\pm$ 0} & 52 $\pm$ 13 & 99.9 $\pm$ 0.1 & 16.2 $\pm$ 41.2 & 124 $\pm$ 21 & 100 $\pm$ 0 & 7.8 $\pm$ 0.44 & \textbf{42 $\pm$ 5} & 87 $\pm$ 19.7 & 16 $\pm$ 17.5 & 2867.7 $\pm$ 230.5\\
10 & 3 & 100 $\pm$ 0 & 10.6 $\pm$ 2.2 & \textbf{43 $\pm$ 15} & 97.4 $\pm$ 3.7 & 296.1 $\pm$ 223.3 & 208 $\pm$ 42 & \textbf{100 $\pm$ 0} & \textbf{10 $\pm$ 0} & 129 $\pm$ 2 & o.o.t. & -- & --\\
20 & 2 & \textbf{100 $\pm$ 0} & \textbf{14.0 $\pm$ 1.8} & \textbf{49 $\pm$ 14 }& 99.9 $\pm$ 0.2 & 196.7 $\pm$ 281.8 & 167 $\pm$ 37 & 100 $\pm$ 0 & 14 $\pm$ 2 & 245 $\pm$ 117 & o.o.t. & -- & --\\
20 & 3 & 99.9 $\pm$ 0.2 & 18.7 $\pm$ 1.7 & \textbf{55 $\pm$ 12} & 69.5 $\pm$ 4.8 & 389.6 $\pm$ 203.4 & 334 $\pm$ 52 & \textbf{100 $\pm$ 0} & \textbf{18.4 $\pm$ 1.1} & 857 $\pm$ 238  & o.o.t. & -- & --\\
30 & 2 & \textbf{100 $\pm$ 0} & \textbf{22.6 $\pm$ 1.3} & \textbf{988 $\pm$ 525} & 99.9 $\pm$ 0 & 77.0 $\pm$ 124.9 & 1216 $\pm$ 122 & o.o.m.& -- & -- & o.o.t & -- & --\\
30 & 3 & \textbf{100 $\pm$ 0} & \textbf{28.0 $\pm$ 1.0} & \textbf{866 $\pm$ 447} & 61.7 $\pm$ 13.8 & 372.9 $\pm$ 171.0 & 2424 $\pm$ 684 & o.o.m.& -- & -- & o.o.t. & -- & --\\
\bottomrule
\end{tabular}
\end{table*}

\begin{table*}[t!]
\caption{Comparison between \textbf{DeepDFA}, \textbf{L* extraction}, \textbf{DFA inductor} and \textbf{DFA-generator} on the \textbf{noisy dataset} created for the \textbf{random DFAs}.}
\label{tbl:comparison_random_dfa_noisy}
\centering
    \setlength{\tabcolsep}{4pt}

\begin{tabular}{lccccccccccccc}
\toprule
\multirow{2}{*}{$|Q|$} & \multirow{2}{*}{$|P|$} & \multicolumn{3}{c}{DeepDFA} & \multicolumn{3}{c}{L* extraction} & \multicolumn{2}{c}{DFA-inductor}  & \multicolumn{2}{c}{DFA-generator} \\
\cmidrule(lr){3-5} \cmidrule(lr){6-8} \cmidrule(lr){9-10} \cmidrule(lr){11-13}
& & Test Acc. & $|\hat{Q}|$ & $\#W$ & Test Acc. & $|\hat{Q}|$ & $\#W$ & Test Acc. & $|\hat{Q}|$ & Test Acc. & $|\hat{Q}|$ &  $\#W$ \\
\midrule
10 & 2 & 93.6 $\pm$ 10.7 & \textbf{11.5 $\pm$ 5.25} & 20200 & \textbf{98.4 $\pm$ 5.0} & 131.5 $\pm$ 225.4 & 62000 & o.o.m.& --  & 84 $\pm$ 14.5 & 15 $\pm$ 18 & \textbf{5050}\\
10 & 3 & \textbf{95.1 $\pm$ 3.6} & \textbf{37.2 $\pm$ 11.1} & \textbf{30200} & 92.0 $\pm$ 8.4 & 290.5 $\pm$ 193.2 & 82600 & o.o.m.& -- & o.o.t. & -- & --\\
20 & 2 & \textbf{99.6 $\pm$ 1.7} & \textbf{14.7 $\pm$ 3.2} & \textbf{20200 }& 98.9 $\pm$ 1.9 & 359.0 $\pm$ 251.2 & 62000 & o.o.m.& -- & o.o.t. & -- & --\\
20 & 3 & \textbf{96.0 $\pm$ 4.1} & \textbf{34.9 $\pm$ 10.1} & \textbf{30200} & 66.2 $\pm$ 4.5 & 420.3 $\pm$ 220.3  & 82600 & o.o.m.& -- & o.o.t. & -- & --\\
30 & 2 & \textbf{99.6 $\pm$ 1.5} & \textbf{22.6 $\pm$ 1.7} & \textbf{80400} & 98.2 $\pm$ 7.4 &238.9 $\pm$ 252.2 & 122200 & o.o.m.& -- & o.o.t. & -- & --\\
30 & 3 & \textbf{98.7 $\pm$ 2.0} & \textbf{55.1 $\pm$ 34.0} & \textbf{120400} & 60.4 $\pm$ 11.3 & 338.7 $\pm$ 173.4 & 326800 & o.o.m.& -- & o.o.t. & -- & --\\
\bottomrule
\end{tabular}
\bigskip
\end{table*}

\paragraph{Dataset} \label{sec:dataset}
For each target DFA, we create a train, a dev, and a test dataset by sampling strings of various lengths and labeling them with the target DFA. 
The training dataset contains strings with a length between 1 and $len_{train}$, the dev set is composed of strings of length $len_{dev}$, and the test set by sequences of length $len_{test}$.
In order to test the model's capability to generalize to longer unseen sequences, for each dataset, we set $len_{train} < len_{dev} < len_{test}$.
To prevent the models from learning degenerate solutions, all the train datasets are nearly perfectly balanced.
We set $len_{train}=30$, $len_{dev}$=60, and $len_{test}$=90 for all the Tomita datasets and the random DFAs datasets with $|Q|<30$.
For the random DFAs with state size of 30, we set $len_{train}$=50, $len_{dev}$=100, and $len_{test}$=150. We report the size of each dataset in the supplementary material.
To test the resiliency of different methods to errors in the training data, we also create a corrupted version of each training dataset by flipping $1\%$ of the labels.

\paragraph{Baselines}
Our approach is a hybrid between a RNN and a DFA. These two types of sequence acceptors are trained with very different methods and present different strengths and weaknesses. In order to better understand the benefits of having a hybrid method, we compare it with one state-of-the-art from the literature on grammar induction, \textbf{DFA inductor} \cite{Zakirzyanov2019EfficientSB}, one state-of-the-art method to extract DFAs from RNNs, \textbf{L* extraction} \cite{weiss18a_queries}, and one state-of-the-art neurosymbolic method for automata learning, \textbf{DFA Generator} \cite{petr2017}. 
Notice we cannot compare with the other neurosymbolic methods cited in Section \ref{sec:related_works}. Indeed, Walke et al. \cite{walke_ltl} learns an LTLf formula from data, LTLf formulas are less expressive than DFAs, since they can capture only \textit{star-free} regular languages \cite{LTLf}, while DFAs can capture both star-free and non-star-free languages. Note that Tomita 3,5 and 6 are non-star-free grammars \cite{transformers_tomita}, and our schedule to produce random DFAs may generate non-star free regular languages. Collery et al. \cite{Collery_rule_induction} learn a set of local and global rules that are ultimately combined, and the relationship between their formalism and that of DFAs is not clear. Regarding the chosen baselines, we remind that DFA-inductor is a SAT-based approach for exact DFA identification. In particular, we use the shared implementation code at \footnote{\href{https://github.com/ctlab/DFA-Inductor-py}{https://github.com/ctlab/DFA-Inductor-py}}, and we test with breadth-first search (BFS) symmetry breaking, shown to be the most effective in the paper.
Instead, L* extraction abstracts a finite state automaton from a pretrained RNN, starting from a predefined discretization of the hidden state space and applying the L* algorithm and abstraction refinement when required.
To test this method, for each language: (i) we train many RNNs of different types, different number of layers and different hidden-state size, (ii) we record the performances on the dev set (iii) we apply L* extraction on the RNN achieving the best dev accuracy. In particular, we use the code in the public repository \footnote{\href{https://github.com/tech-srl/lstar_extraction}{https://github.com/tech-srl/lstar\_extraction}} with 10 as the initial split depth, and the starting examples set composed of the shortest positive string and the shortest negative string in the train set, as suggested in the paper.
DFA Generator trains directly a neural network shaped as a DFA, as our method. The algorithm needs as hyperparameter the maximum number of states $\hat{Q_{max}}$. We train with $\hat{Q_{max}}$ of different sizes (the same used for our method when it is possible), and we use the implementation at the public repository \footnote{\href{https://github.com/pgrachev/Automaton-Generator}{https://github.com/pgrachev/Automaton-Generator}}.
For each approach, we report the accuracy obtained by querying the final DFA on the test set, the number of states of the predicted DFA $|\hat{Q}|$, the execution time (in seconds), and, when applicable, the number of weights of the model.

\begin{figure*}[t]
\centering
\subfigure[]{\includegraphics[width=0.245\textwidth]{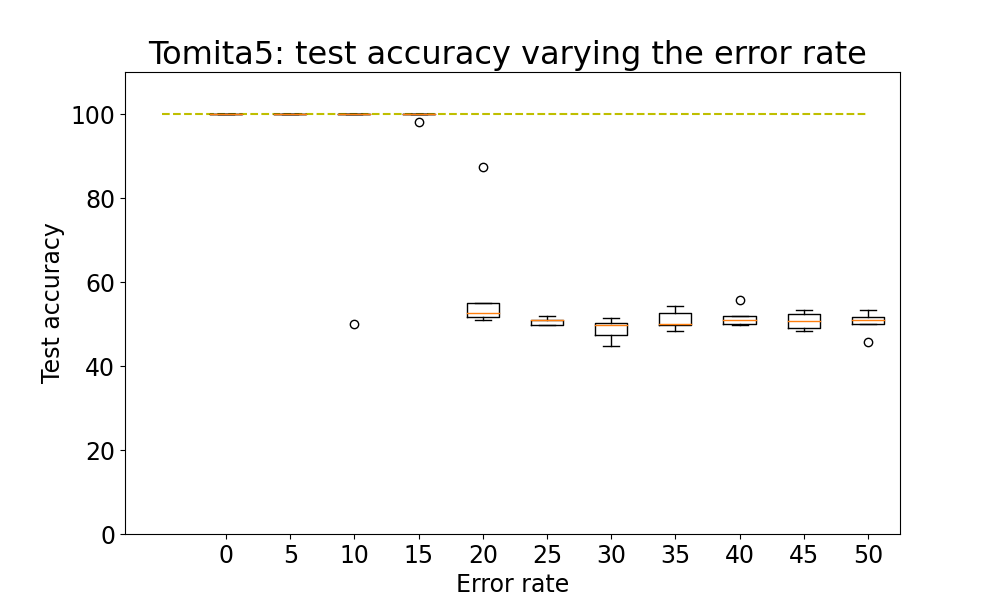}}
\subfigure[]{\includegraphics[width=0.245\textwidth]{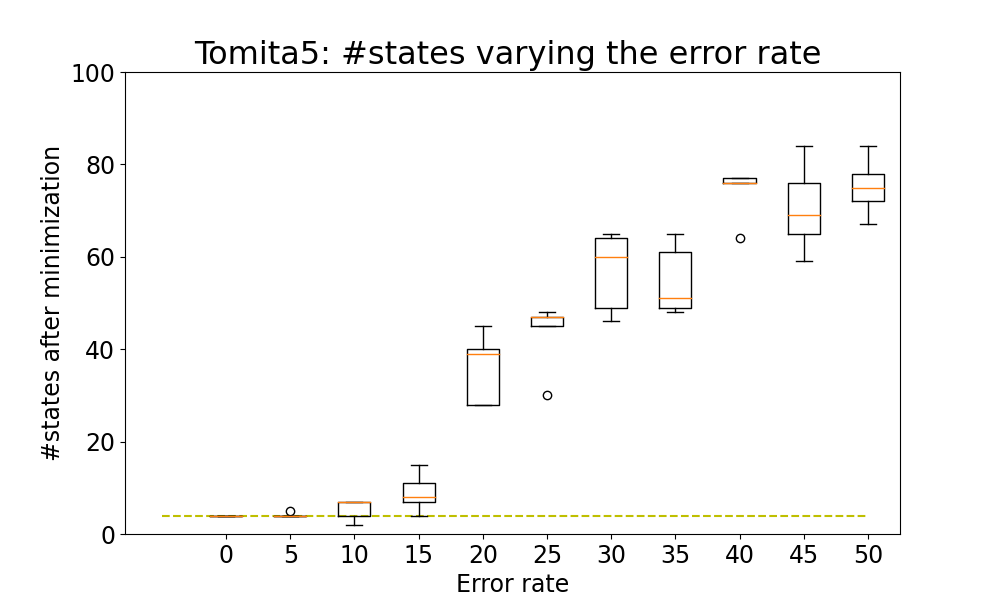}}
\subfigure[]{\includegraphics[width=0.245\textwidth]{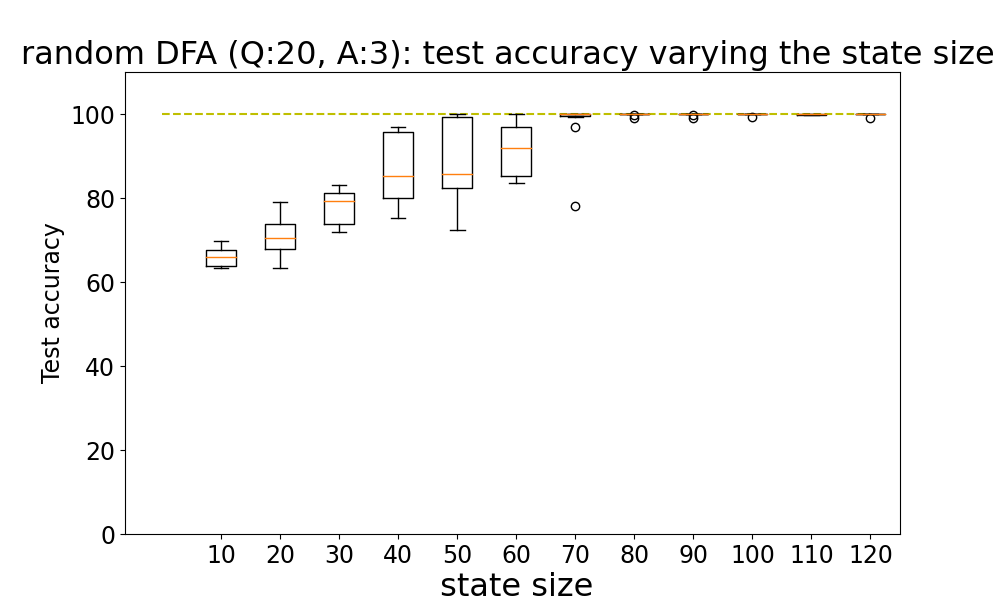}}
\subfigure[]{\includegraphics[width=0.245\textwidth]{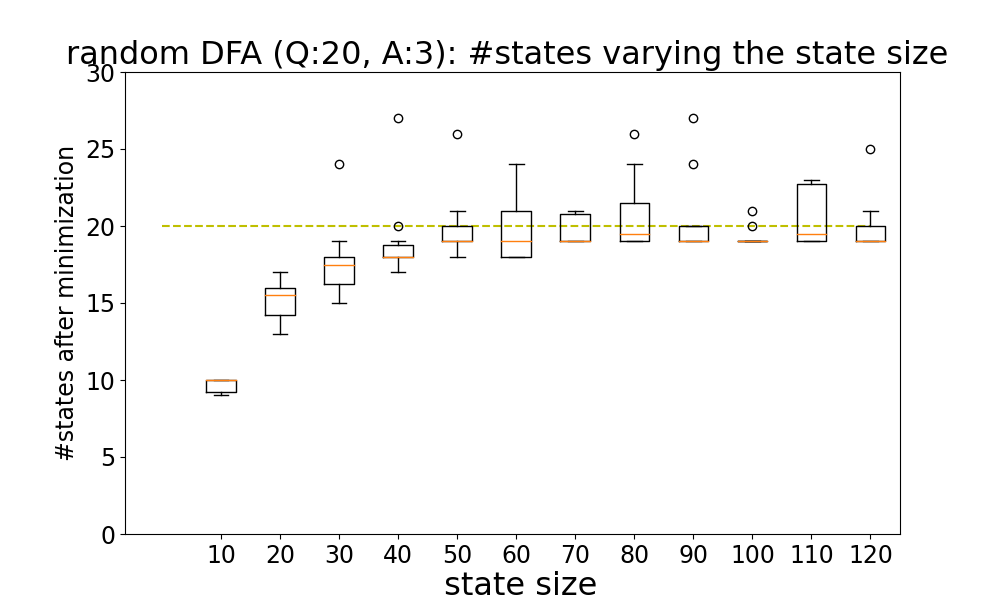}}
\caption{Ablation study 1: \textbf{results on Tomita 5 with different error rates in the training dataset}. a) Test accuracy. b) Number of states of predicted DFAs. Ablation study 2: \textbf{results a random DFA of size 20 and alphabet size 3 varying the state size hyperparameter}. c) Test accuracy. d) Predicted number of states. Each box represents 10 experiments performed with different random seeds. The box extends from the lower to upper quartile values of the data, with the line at the median. The whiskers extend from the box to show the range of the data. Flier points represent outliers.}
\label{fig:ablation_study}

\end{figure*}
\paragraph{Training Details}
We train all the neural networks on an Nvidia GPU GeForce GTX 1650 Ti, with a learning rate of 0.01 for DeepDFA and of 0.001 for the RNNs, until training loss convergence, for a maximum of 200 epochs. In all the experiments we use $\lambda=0.999$ and minimum temperature $= 10^{-5}$.
All the experiments of DFA-inductor were performed on a Intel Core i7-10750H CPU, without any other process running at the same time.

\subsection{Results}
\paragraph{Results on Tomita Languages}
For each Tomita language, we report the results averaged over 5 experiments with different random seeds for L* extraction, DFA-generator and DeepDFA, and the results of one application of DFA-inductor, since the latter has a deterministic behavior. To explore the influence of hyperparameter choices on DeepDFA, DFA-generator and L* extraction, we test 3 different architectures for DeepDFA and DFA-generator and 6 different RNNs for L*. In particular, we test DeepDFA and DFA-generator with the maximum number of states equal to 10, 30, and 100 (denoting the models as DeepDFA($|\hat{Q}_{max}|$) and DFAGen(($|\hat{Q}_{max}|$)). We test L* extraction on both LSTMs and GRUs, for each type of RNN we test three architectures (denoted here as RNN(\#layers, \#hidden neurons per layer) ), namely: RNN(1,30), RNN(1,100), RNN(2,100). 
For space reasons, we report in the supplementary material the results obtained for each model tested and here only the results of the best model for each approach. We select he best model per approach as the more accurate on the development set. In cases where two or more models achieve the same highest accuracy on the development set, we prioritize selecting the model with the smallest number of weights among them. Results for the Tomita benchmark are reported in Table \ref{tbl:comparison_tomita}. 
For space reasons we do not report the execution time for the Tomita languages. However each run last less than a minute across the all 7 languages for all the models tested for all the approaches, except for DFA-generator. This proved to be the slowest of all the methods tested. In particular we set a threshold time of one hour for a run and we denote with `o.o.t' (out of time) when a method exceed the threshold time. DFAGen(10) takes less than a minute only for Tomita 1, 2 and 3, and never exceed the timeout. DFAGen(30) execution time is less than a minute only for Tomita 1 and 2 and exceed the maximum time for Tomita 5 and 7, while DFAGen(100) always goes out of maximum time.
Remarkably, both our approach and DFA-inductor achieve perfect performances on all the the 7 languages, namely a test accuracy of 100\% and a predicted number of states equal to the target. The same is achieved by L* extraction, for all the grammars except Tomita 6, for which it predicts extremely oversized DFAs with a mean of 170 states. DFA-generator instead can produce optimal results only for Tomita 6 and has lower performances on all the other languages.
To evaluate how noise in the training data affects the performance of the four methods, we replicate the experiment using training datasets with 1\% label noise. The results are presented in Table \ref{tbl:comparison_tomita_noisy}.
Results show that DFA-inductor is completely unable to handle errors in the training data. In fact, with only 1\% of errors, the process is killed for exceeding the CPU capacity before finding a DFA consistent with training examples. In the table, these instances are marked with an `o.o.m' (out of memory).
The method crashes in all the Tomita languages except the first two, for which it loses around 5\% of accuracy and overestimates the state space. As expected, the other methods based on neural networks exhibits greater robustness to training set errors. DFA-generator presents non optimal performances similar to the case without noise. L* extraction overestimates the number of states for Tomita3 and 6. In contrast, our approach remains unaffected by minor errors in the training dataset and consistently maintains the top performance levels achieved with the error-free dataset. As an ablation study, we subjected DeepDFA to different error rates (ranging from 0 to 50\%) on Tomita 5. Each noise configuration was tested across five experiments, with $|\hat{Q}_{max}|$ set to 200. The results in Figure \ref{fig:ablation_study}(a-b) show that our model can tolerate larger error rates of up to 15\% achieving 100\% test accuracy and a state count similar to the target one.
\paragraph{Results on randomly generated DFAs}
We generate random DFAs with varying state and alphabet sizes using the approach outlined in the `Target DFAs' paragraph. For each randomly generated DFA, we conduct 10 tests for each stochastic methods (DeepDFA, L* extraction and DFA-generator) and one test with DFA-inductor. For this benchmark we set a larger threshold time of two hours for each run. We configure DeepDFA with a state size $|\hat{Q}{max}|$ of 100 for random DFAs with state sizes of 10 and 20, and $|\hat{Q}{max}|$ of 200 when the desired number of states is 30. When testing L* extraction, we train both LSTM and GRU RNNs of one and two layers, using the same hidden state sizes used for DeepDFA. We initially tested DFA-generator with $|\hat{Q}{\text{max}}|$ set to 100, only to find that it always exceeded the time limit. Consequently, we experimented with a reduced $|\hat{Q}{\text{max}}|$ value of 50. Despite this adjustment, the method continued to go out-of-time for the majority of languages. This happens for a difficulty in reducing the loss function, causing the method to remain stalled.
However, this confirms the scalability challenges previously acknowledged by the authors of the method in \cite{petr2017}, highlighting the method's struggle to converge when dealing with automata larger than five states. Our comparison of the four methods is presented in Table \ref{tbl:comparison_random_dfa}. For the stochastic methods, we provide the average performance over the best 5 out of 10 runs.
DFA-inductor is able to reach top test accuracy unless it crashes due to exceeding CPU capacity; this again happens for the biggest-size DFAs of size 30. Our approach competes favorably with DFA-inductor for target DFAs smaller than size 30, and it excels in cases where DFA-inductor falls short. Compared to L* extraction and DFA-generator, our method consistently exhibits higher accuracy. Notably, L* extraction tends to significantly overestimate the number of states, while DeepDFA predicts DFAs with state counts comparable to the ground truth size. Moreover, our method proves to be the fastest among the four, except in a single case.
We repeat the experiments on a corrupted version of the training datasets, where 1\% of labels are erroneous. The results are presented in Table \ref{tbl:comparison_random_dfa_noisy}, showing the average performance over 10 trials. In this scenario, DFA-inductor is unable to provide a solution even for the smallest size DFAs. DeepDFA, L* extraction and DFA-generator (when it can converge) display only minor drops in performance, demonstrating resilience to noise. DeepDFA outperforms both the other stochastic methods in terms of test accuracy and state count prediction, and almost always employs the fewest number of parameters. 
\textbf{In summary}, the main issues encountered in competitor methods are:
(i) scalability problems with DFA-inductor and DFA-generator, often resulting in out-of-memory or out-of-time errors when handling large DFAs.
(ii) very low resilience to errors exhibited by DFA-inductor.
(iii) tendency to significantly overestimate the automaton state space, possibly due to overfitting, observed in L* extraction.
In contrast, our method consistently delivers the most precise results in terms of test accuracy and number of states. It achieves this without encountering any out-of-time or out-of-memory issues, while also demonstrating a competitive number of parameters and execution times across all experiments.
\begin{figure*}[t!]
    \centering
    \subfigure[Tomita1]{\includegraphics[width=0.23\textwidth]{ 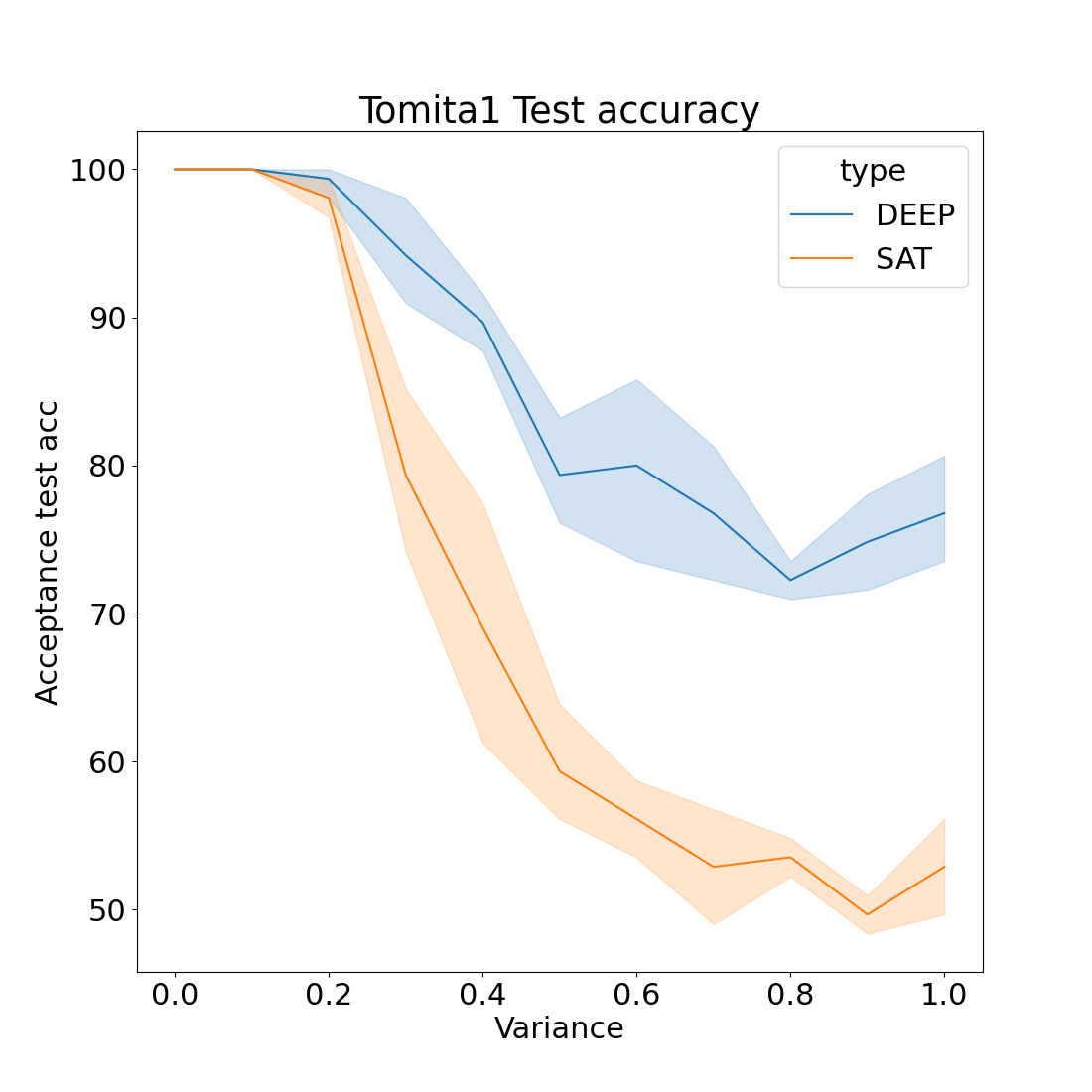}
    }
    \subfigure[Tomita2]{\includegraphics[width=0.23\textwidth]{ 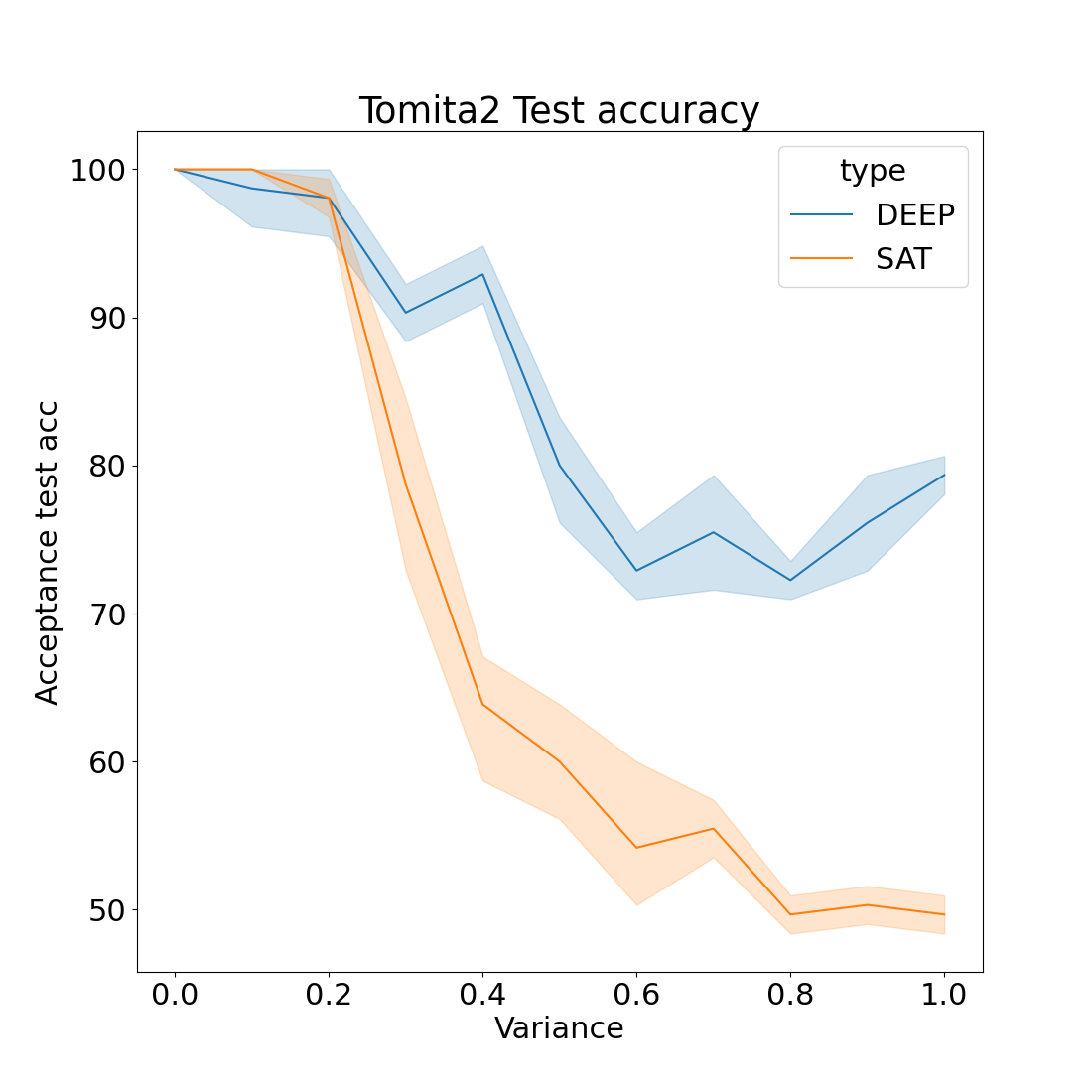}
    }
    \subfigure[Tomita3]{\includegraphics[width=0.23\textwidth]{ 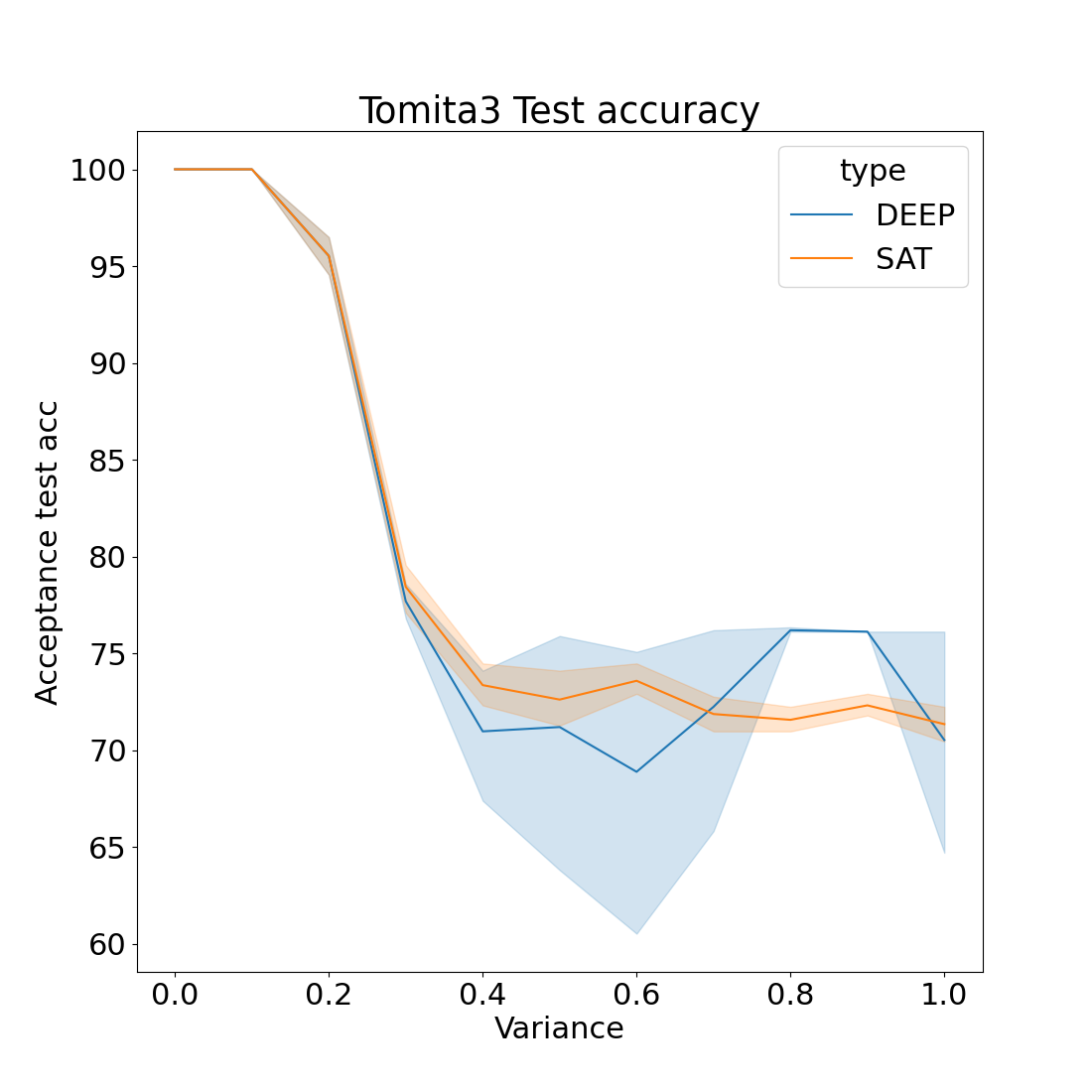}
    }
    \subfigure[Tomita4]{\includegraphics[width=0.23\textwidth]{ 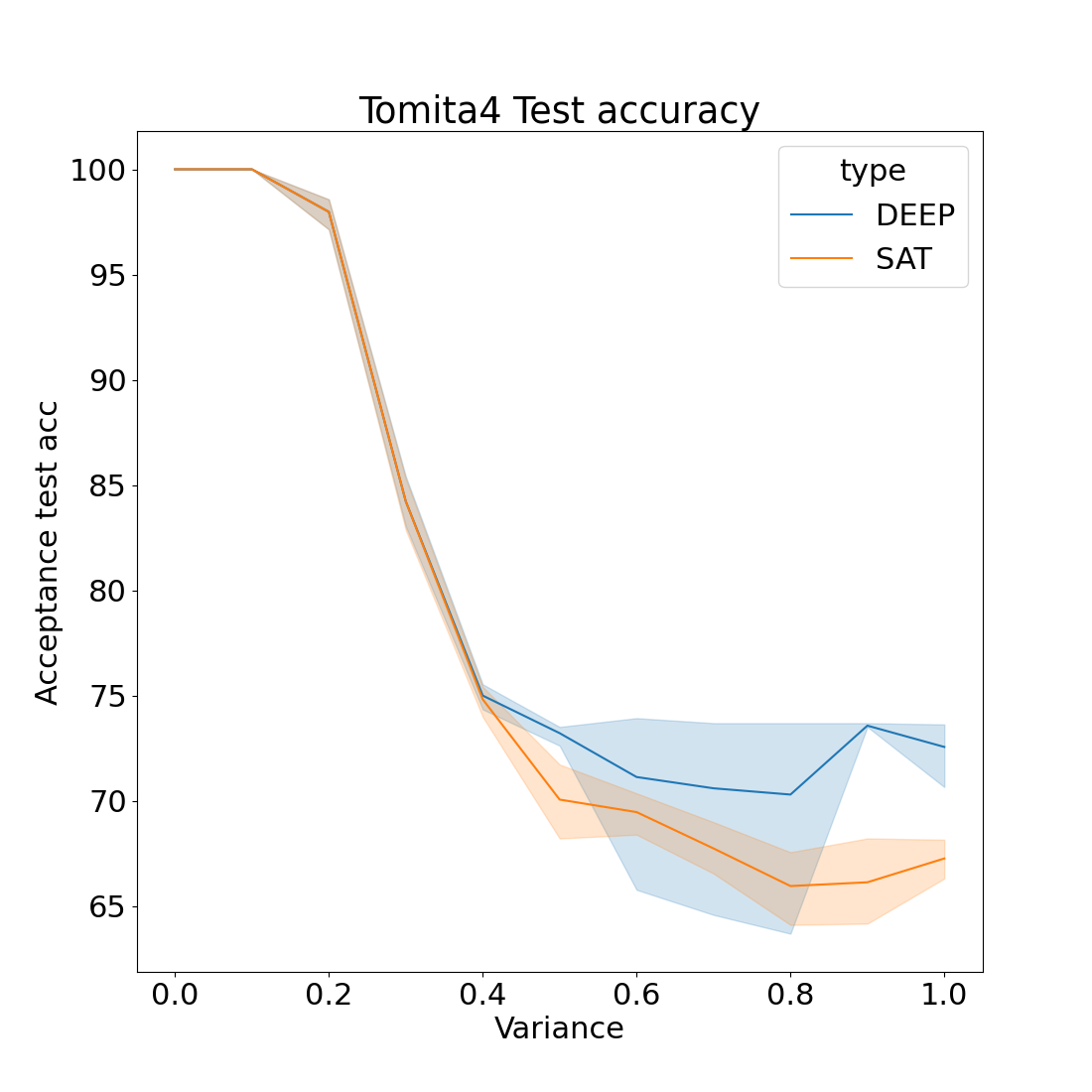}
    }
    \subfigure[Tomita5]{\includegraphics[width=0.23\textwidth]{ 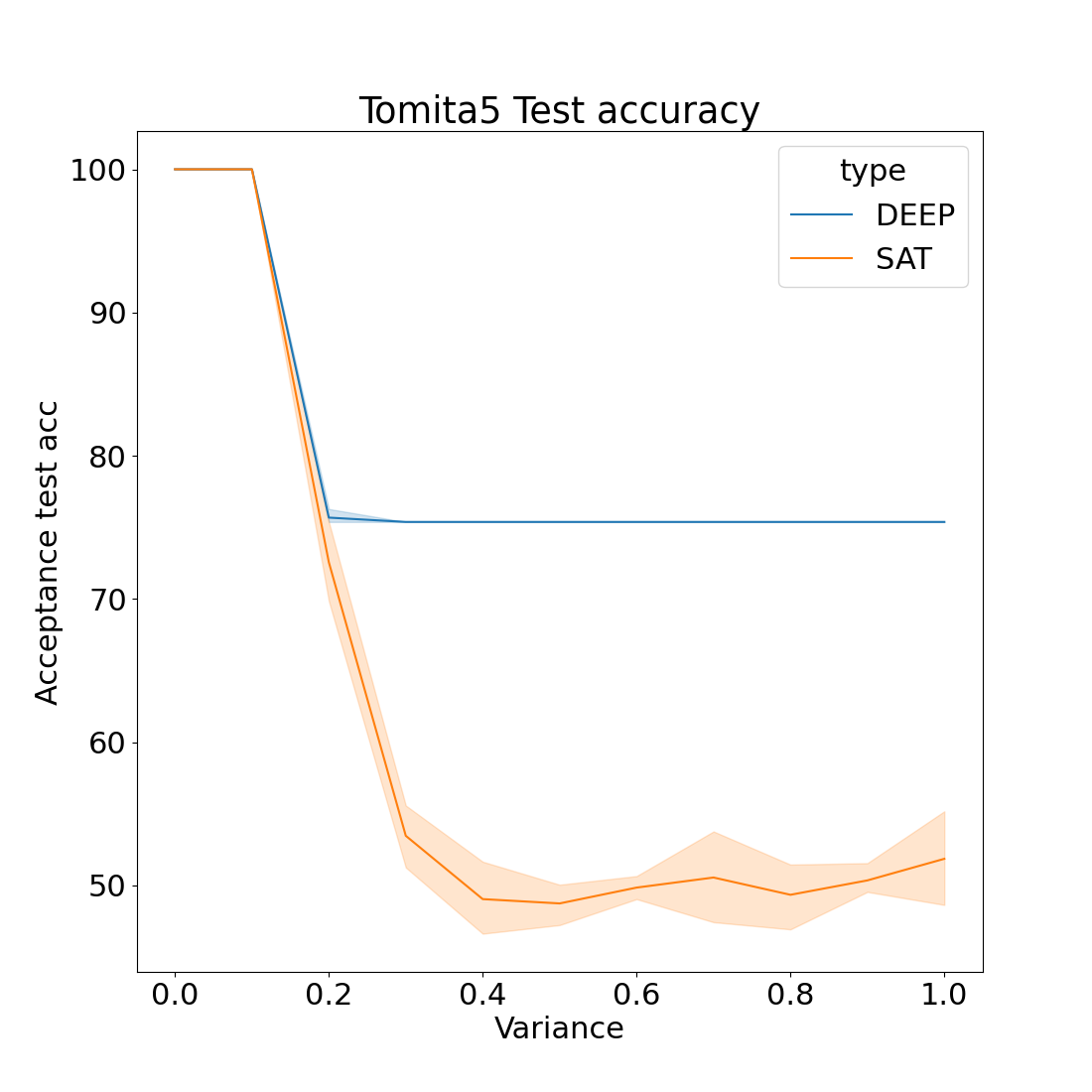}
    }
    \subfigure[Tomita6]{\includegraphics[width=0.23\textwidth]{ 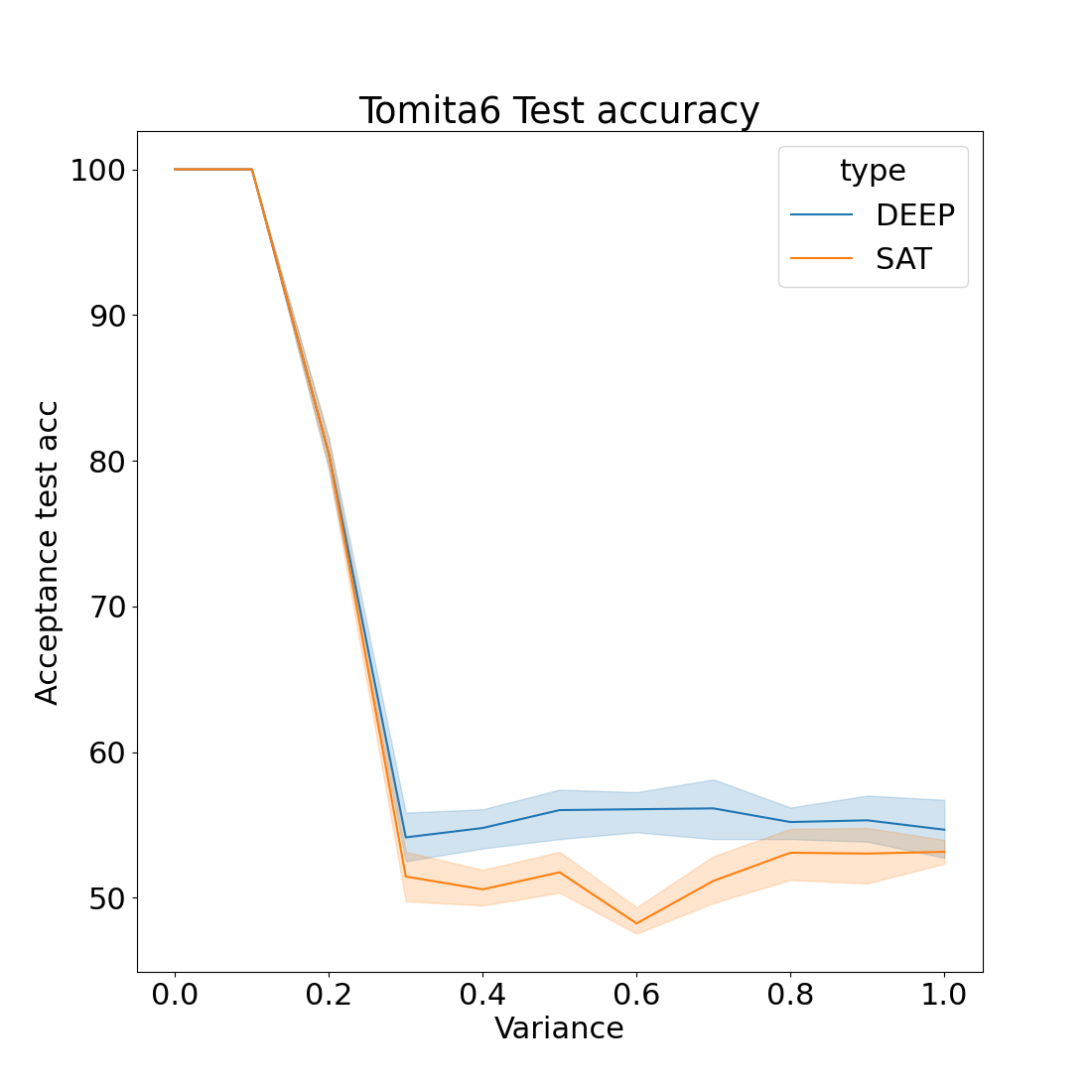}
    }
    \subfigure[Tomita7]{\includegraphics[width=0.23\textwidth]{ 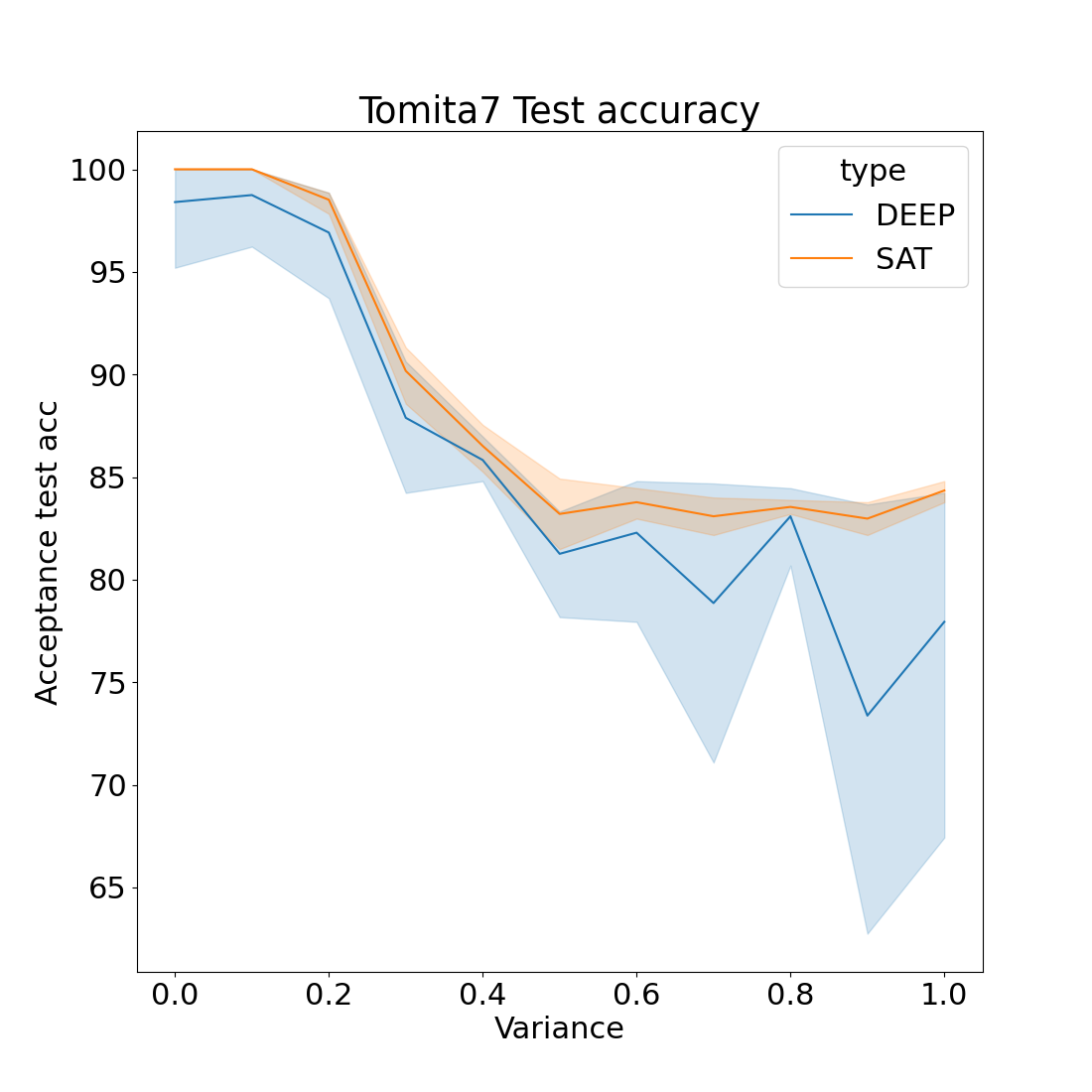}
    }
    
    \caption{Learning DFA from traces composed of imperfectly grounded symbols: test accuracy on the 7 Tomita languages. In blue the results obtained with the extension of DeepDFA to probabilistic symbols, in orange results of DFA-inductor.}
    \label{fig:tomita_uncertain_symbols}
\end{figure*}
\paragraph{Learning DFAs from traces of imperfectly grounded symbols}
Automata learning typically takes a set of symbolic traces as input, which are sequences of propositional symbols grounded in a perfectly binary manner. However, in many domains, such for example Reinforcement Learning \cite{noisy_symbols_2022_shila, Visual_reward_machine}, the observable sequences come with uncertainty in their grounding. Specifically, at each step of the trace, we have a set of probabilistic beliefs over the symbol set instead of a binary interpretation. To the best of our knowledge, this is the first approach to automata learning that can accommodate such `probabilistic grounding' of symbols in the traces.
To evaluate the capability of our framework in learning DFA specifications from sequences of imperfectly grounded symbols, we utilized a modified version of the dataset for Tomita Languages, where the one-hot representation of input symbols in the training traces is corrupted with Gaussian noise having a mean of zero and variable variance. We compared the extension of DeepDFA to handle probabilistic symbols (Equation \ref{eq:deepDFA_uncertain_sym}) with DFA-inductor. Notably, boolean logic induction methods, such as DFA-inductor, cannot handle probabilistic truth values for the input symbols. Therefore, we discretized the corrupted inputs to the nearest one-hot vector before passing them to the method.
Figure \ref{fig:tomita_uncertain_symbols} shows that DeepDFA exhibits a greater resilience to noise in the input symbols than its competitor and its performance decrease smother as we increase the noise intensity.
\paragraph{Ablation study: the effect of changing the state space size} \label{sec:ablationstudy}
RNNs have many hyperparameters and design choices that can affect performance: the model type, the number of layers, the number of features per layer, and more. 
By contrast, our recurrent neural model is a simplified structure with only one hyperparameter: the hidden state size. In this section, we discuss this hyperparameter choice. 
Figure \ref{fig:ablation_study}(c-d) shows the effect on the test accuracy and the predicted number of states of tuning this hyperparameter.
We conducted experiments using a random DFA with a size of 20 and an alphabet size of 3. The hidden state size was varied within the range of 10 to 120.
Results show that our model demands a hidden state size surpassing the actual number of states. 
The best performances for the random DFA of size 20, are achieved for $\hat{Q}_{max} \geq 70$—exceeding its actual state count by 3.5 times. Remarkably, the model maintains its robust performance even when the state size is markedly overestimated. Even at the largest state size, test accuracy remains commendably high, and the number of states inferred remains proximate to the actual count. This suggests that the model evades overfitting; despite being dimensioned to represent a substantially greater number of states, it leverages only a subset of them.

\section{Conclusions and Future Work}
In conclusion, we propose DeepDFA: a hybrid between a DFA and a recurrent neural network, which can be trained from samples with backpropagation as usual deep learning models, but that is completely interpretable, as a DFA, after training. Our approach takes the best from the two worlds: grammar induction on one side and recurrent neural networks on the other side. It uses fewer weights and only requires setting one hyperparameter compared to recurrent neural nets. At the same time, it can tolerate errors in the training labels and noise in the input symbols, and can be applied to large target DFAs, differently from exact methods.
DeepDFA can find applications in various scenarios. In this paper, we present the framework in a general context, without specifying any particular application. However, we are keen to apply it in the area of non-Markovian Deep Reinforcement Learning \cite{apll_1, brafman_ecai_2023, noisy_symbols_2022_shila}.
Furthermore, we are developing a multi-layered version of DeepDFA, having the potential to expand its capabilities to even more intricate regular languages without compromising the explainability, which we leave for future research.

\begin{ack}
This work has been partially supported by PNRR MUR project PE0000013-FAIR. 
\end{ack}




\appendix

\section{Experiments}
\subsection{Datasets}
We report the statistics about the datasets used for our experiments on the Tomita languages and the random DFAs in the Tables \ref{tbl:dataset_tomita} \ref{tbl:dataset_random_dfa} respectively. The tables report the number of traces composing the train, dev, and test dataset and the lengths of the traces in each dataset. Let $len_{train}$ be the maximum length of train traces. Note that we always choose the length of dev traces as $2*len_{train}$ and the length of test traces as $3*len_{train}$, to test the models' capability to generalize to longer unseen traces. Each train dataset is nearly perfectly balanced between positive and negative examples.
\subsection{Training details}
    
We train all the neural networks on an Nvidia GPU GeForce GTX 1650 Ti, with a learning rate of 0.01 for DeepDFA and of 0.001 for the RNNs, until training loss convergence, for a maximum of 200 epochs. In all the experiments we use $\lambda=0.999$ and minimum temperature $= 10^{-5}$.
All the experiments of DFA-inductor were performed on a Intel Core i7-10750H CPU, without any other process running at the same time.

\begin{table}[t!]
    \centering
    \caption{Statistics on the datasets for the Tomita experiments.}
    \label{tbl:dataset_tomita}
    \centering
    \setlength{\tabcolsep}{2pt}
    \begin{tabular}{ccccccc}
        \toprule
        \multirow{2}{*}{Tomita} & \multicolumn{2}{c}{Train Set} & \multicolumn{2}{c}{Dev Set} & \multicolumn{2}{c}{Test Set} \\
        \cmidrule(lr){2-3} \cmidrule(lr){4-5} \cmidrule(lr){6-7}
        & Size & Length & Size & Length & Size & Length \\
        \midrule
        1 & 325 & 1-30 & 20 & 60 & 20 & 90 \\
        2 & 315 & 1-30 & 20 & 60 & 20 & 90 \\
        3 & 2701 & 1-30 & 20 & 60 & 20 & 90 \\
        4 & 3373 & 1-30 & 40 & 60 & 20 & 90 \\
        5 & 1991 & 1-30 & 300 & 60 & 300 & 90 \\
        6 & 3423 & 1-30 & 300 & 60 & 300 & 90 \\
        7 & 1741 & 1-30 & 20 & 60 & 20 & 90 \\
        \bottomrule
    \end{tabular}
\end{table}

\begin{table}[]
    \centering
    \caption{Statistics on the datasets used for the random DFA experiments.}
    \label{tbl:dataset_random_dfa}
    \centering
    \setlength{\tabcolsep}{2pt}
    \begin{tabular}{cccccccc}
        \toprule
        \multirow{2}{*}{S} & \multirow{2}{*}{A} & \multicolumn{2}{c}{Train Set} & \multicolumn{2}{c}{Dev Set} & \multicolumn{2}{c}{Test Set} \\
        \cmidrule(lr){3-4} \cmidrule(lr){5-6} \cmidrule(lr){7-8}
        & & Size & Length & Size & Length & Size & Length \\
        \midrule
        10 & 2 & 3341.4 $\pm$ 159.95 & 1-30 & 150  & 60 & 150  & 90 \\
        10 & 3 & 4196.6 $\pm$ 31.32 & 1-30 & 150  & 60 & 150  & 90 \\
        20 & 2 & 3434.6 $\pm$ 46.52 & 1-30 & 150  & 60 & 150  & 90 \\
        20 & 3 & 4213.0 $\pm$ 16.73 & 1-30 & 150  & 60 & 150  & 90 \\
        30 & 2 & 12804.6 $\pm$ 11.67 & 1-50 & 150  & 100 & 150  & 150 \\
        30 & 3 & 13556.8 $\pm$ 29.01 & 1-50 & 150  & 100 & 150  & 150 \\
        \bottomrule
    \end{tabular}
\end{table}

\begin{table*}[t!]
\centering
\caption{\textbf{L* extraction} on the \textbf{Tomita} Languages with different RNN architectures. We highlight in bold the models that have obtained perfect performances for T1-T4. Since no model can reach perfect results on T5 and T6, for these languages, we highlight in bold the model reaching the best RNN accuracy on the dev set. All the RNN models achieve perfect results on T7; for this and space reasons, we omit the results of this language from the table.}
\label{tbl:L_star_tomita}
\begin{tabular}{cccccccccc}
\toprule
\multirow{2}{*}{Model} & \multicolumn{4}{c}{T1 ($|Q| = 2$)} & \multicolumn{4}{c}{T2 ($|Q| = 3$)} \\
\cmidrule(lr){2-5} \cmidrule(lr){6-9}
& Train & Dev & Test & $|\hat{Q}|$ & Train & Dev & Test & $|\hat{Q}|$ \\
\midrule
LSTM(1,30) & 100 $\pm$ 0 & 100 $\pm$ 0 & --- & --- & 100 $\pm$ 0 & 100 $\pm$ 0 & --- & --- \\
DFA extracted & 99.3 $\pm$ 1.3 & 100 $\pm$ 0 & 100 $\pm$ 0 & 2.2 $\pm$ 0.4 & 99.4 $\pm$ 0.7 & 100 $\pm$ 0 & 100 $\pm$ 0 & 2.6 $\pm$ 0.5 \\
\midrule
LSTM(1,100) & 100 $\pm$ 0 & 100 $\pm$ 0 & --- & --- & \textbf{100 $\pm$ 0} & \textbf{100 $\pm$ 0} & --- & --- \\
DFA extracted & 99.8 $\pm$ 0.2 & 100 $\pm$ 0 & 100 $\pm$ 0 & 2.6 $\pm$ 1.3 & \textbf{100 $\pm$ 0} & \textbf{100 $\pm$ 0} & \textbf{100 $\pm$ 0} & \textbf{3 $\pm$ 0} \\
\midrule
LSTM(2,100) & \textbf{100 $\pm$ 0} & \textbf{100 $\pm$ 0} & --- & --- & 100 $\pm$ 0 & 100 $\pm$ 0 & --- & --- \\
DFA extracted & \textbf{100 $\pm$ 0} & \textbf{100 $\pm$ 0} & \textbf{100 $\pm$ 0} & \textbf{2$\pm$0} & 98.73 $\pm$ 0 & 100 $\pm$ 0 & 100 $\pm$ 0 & 2 $\pm$ 0 \\
\midrule
GRU(1,30) & \textbf{100 $\pm$ 0} & \textbf{100 $\pm$ 0} & --- & --- & \textbf{100$\pm$ 0} & \textbf{100 $\pm$ 0} & --- & --- \\
DFA extracted & \textbf{100 $\pm$ 0} & \textbf{100 $\pm$ 0} & \textbf{100 $\pm$ 0} & \textbf{2 $\pm$ 0} & \textbf{100 $\pm$ 0} & \textbf{100 $\pm$ 0} & \textbf{100 $\pm$ 0} & \textbf{3 $\pm$ 0} \\
\midrule
GRU(1,100) & 100 $\pm$ 0 & 100 $\pm$ 0 & --- & --- & 100 $\pm$ 0 & 100 $\pm$ 0 & --- & --- \\
DFA extracted & 99.3 $\pm$ 1.3 & 100 $\pm$ 0 & 100 $\pm$ 0 & 2.2 $\pm$ 0.4 & 100 $\pm$ 0 & 100 $\pm$ 0 & 100 $\pm$ 0 & 4.4 $\pm$ 3.1 \\
\midrule
GRU(2,100) & \textbf{100 $\pm$ 0} & \textbf{100 $\pm$ 0} & --- & --- & 100 $\pm$ 0 & 100 $\pm$ 0 & --- & --- \\
DFA extracted & \textbf{100 $\pm$ 0} & \textbf{100 $\pm$ 0} & \textbf{100 $\pm$ 0} & \textbf{2 $\pm$ 0} & 98.7 $\pm$ 0 & 100 $\pm$ 0 & 100 $\pm$ 0 & 2 $\pm$ 0 \\
\bottomrule
\end{tabular}
\begin{tabular}{cccccccccc}
\toprule
\multirow{2}{*}{Model} & \multicolumn{4}{c}{T3 ($|Q| = 5$)} & \multicolumn{4}{c}{T4 ($|Q| = 4$)} \\
\cmidrule(lr){2-5} \cmidrule(lr){6-9}
& Train & Dev & Test & $|\hat{Q}|$ & Train & Dev & Test & $|\hat{Q}|$ \\
\midrule
LSTM(1,30) & \textbf{100 $\pm$ 0} & \textbf{100 $\pm$ 0} & --- & --- & \textbf{100 $\pm$ 0} & \textbf{100 $\pm$ 0} & --- & --- \\
DFA extracted & \textbf{100 $\pm$ 0} & \textbf{100 $\pm$ 0} & \textbf{100 $\pm$ 0} & \textbf{5 $\pm$ 0} & \textbf{100 $\pm$ 0} & \textbf{100 $\pm$ 0} & \textbf{100 $\pm$ 0} & \textbf{4 $\pm$ 0} \\
\midrule
LSTM(1,100) & \textbf{100 $\pm$ 0} & \textbf{100 $\pm$ 0} & --- & --- & \textbf{100 $\pm$ 0} & \textbf{100 $\pm$ 0} & --- & --- \\
DFA extracted & \textbf{100 $\pm$ 0} & \textbf{100 $\pm$ 0} & \textbf{100 $\pm$ 0} & \textbf{5 $\pm$ 0} & \textbf{100 $\pm$ 0} & \textbf{100 $\pm$ 0} & \textbf{100 $\pm$ 0} & \textbf{4 $\pm$ 0} \\
\midrule
LSTM(2,100) & \textbf{100 $\pm$ 0} & \textbf{100 $\pm$ 0} & --- & --- & \textbf{100 $\pm$ 0} & \textbf{100 $\pm$ 0} & --- & --- \\
DFA extracted & \textbf{100 $\pm$ 0} & \textbf{100 $\pm$ 0} & \textbf{100 $\pm$ 0} & \textbf{5 $\pm$ 0} & \textbf{100 $\pm$ 0} & \textbf{100 $\pm$ 0} & \textbf{100 $\pm$ 0} & \textbf{4 $\pm$ 0} \\
\midrule
GRU(1,30) & \textbf{100 $\pm$ 0} & \textbf{100 $\pm$ 0} & --- & --- & \textbf{100 $\pm$ 0} & \textbf{100 $\pm$ 0} & --- & --- \\
DFA extracted & \textbf{100 $\pm$ 0} & \textbf{100 $\pm$ 0} & \textbf{100 $\pm$ 0} & \textbf{5 $\pm$ 0} & \textbf{100 $\pm$ 0} & \textbf{100 $\pm$ 0} & \textbf{100 $\pm$ 0} & \textbf{4 $\pm$ 0} \\
\midrule
GRU(1,100) & 100 $\pm$ 0 & 100 $\pm$ 0 & --- & --- & \textbf{100 $\pm$ 0} & \textbf{100 $\pm$ 0} & --- & --- \\
DFA extracted & 99.9 $\pm$ 0 & 100 $\pm$ 0 & 100 $\pm$ 0 & 285.4 $\pm$ 384 & \textbf{100 $\pm$ 0} & \textbf{100 $\pm$ 0} & \textbf{100 $\pm$ 0} & \textbf{4 $\pm$ 0} \\
\midrule
GRU(2,100) & \textbf{100 $\pm$ 0} & \textbf{100 $\pm$ 0} & --- & --- & \textbf{100 $\pm$ 0} & \textbf{100 $\pm$ 0} & --- & --- \\
DFA extracted & \textbf{100 $\pm$ 0} & \textbf{100 $\pm$ 0} & \textbf{100 $\pm$ 0} & \textbf{5 $\pm$ 0} & \textbf{100 $\pm$ 0} & \textbf{100 $\pm$ 0} & \textbf{100 $\pm$ 0} & \textbf{4 $\pm$ 0} \\
\bottomrule
\end{tabular}
\centering
\setlength{\tabcolsep}{4pt} 

\begin{tabular}{cccccccccc}
\toprule
\multirow{2}{*}{Model} & \multicolumn{4}{c}{T5 ($|Q| = 4$)} & \multicolumn{4}{c}{T6 ($|Q| = 3$)} \\
\cmidrule(lr){2-5} \cmidrule(lr){6-9}
& Train & Dev & Test & $|\hat{Q}|$ & Train & Dev & Test & $|\hat{Q}|$ \\
\midrule
LSTM(1,30) & 76.4 $\pm$ 0.2 & 49.8 $\pm$ 0.7 & --- & --- & 87.2 $\pm$ 0.7 & 56.6 $\pm$ 3.3 & --- & --- \\
DFA extracted & 53.7 $\pm$ 0 & 50 $\pm$ 0 & 50 $\pm$ 0 & 2 $\pm$ 0 & 81.7 $\pm$ 0.6 & 56.4 $\pm$ 3.7 & 51.9 $\pm$ 2.5 & 621.8 $\pm$ 243.5 \\
\midrule
LSTM(1,100) & 78.8 $\pm$ 1.3 & 51.87 $\pm$ 0.51 & --- & --- & \textbf{99 $\pm$ 2} & \textbf{92.6 $\pm$ 16.4} & --- & --- \\
DFA extracted & 54.5 $\pm$ 0.8 & 50.6 $\pm$ 0.9 & 48.7 $\pm$ 1.7 & 77.6 $\pm$ 156.2 & \textbf{98.7 $\pm$ 2.8} & \textbf{92.8 $\pm$ 16.1} & \textbf{91.1 $\pm$ 19.8} & \textbf{170.6 $\pm$ 374.7} \\
\midrule
LSTM(2,100) & 79 $\pm$ 1 & 50.6 $\pm$ 1.2 & --- & --- & 95.9 $\pm$ 2.8 & 66.6 $\pm$ 8.3 & --- & --- \\
DFA extracted & 54.5 $\pm$ 0.8 & 50.4 $\pm$ 0.8 & 49.6 $\pm$ 1.6 & 110.4 $\pm$ 213.1 & 96.2 $\pm$ 1.5 & 70.3 $\pm$ 8 & 65.1 $\pm$ 6.6 & 233.8 $\pm$ 208.4 \\
\midrule
GRU(1,30) & 78.4 $\pm$ 1.1 & 51 $\pm$ 1.3 & --- & --- & 82.8 $\pm$ 3.7 & 54.7 $\pm$ 3.7 & --- & --- \\
DFA extracted & 54.5 $\pm$ 0.8 & 51.2 $\pm$ 1.7 & 47.9 $\pm$ 1.6 & 307.6 $\pm$ 295.5 & 79.9 $\pm$ 12 & 60.9 $\pm$ 21.9 & 60.9 $\pm$ 21.9 & 408.2 $\pm$ 245.1 \\
\midrule
GRU(1,100) & 100 $\pm$ 0 & 98.9 $\pm$ 1.2 & --- & --- & 98.7 $\pm$ 1.3 & 86.5 $\pm$ 13.1 & --- & --- \\
DFA extracted & 100 $\pm$ 0 & 100 $\pm$ 0 & 100 $\pm$ 0 & 4 $\pm$ 0 & 98.6 $\pm$ 1.9 & 89.7 $\pm$ 14.1 & 85.7 $\pm$ 19.5 & 198.6 $\pm$ 371.4 \\
\midrule
GRU(2,100) & \textbf{100 $\pm$ 0} & \textbf{99.9 $\pm$ 0.1} & --- & --- & 97.3 $\pm$ 0.99 & 62 $\pm$ 4.2 & --- & --- \\
DFA extracted & \textbf{100 $\pm$ 0} & \textbf{100 $\pm$ 0} & \textbf{100 $\pm$ 0} & \textbf{4 $\pm$ 0} & 95.9 $\pm$ 2.7 & 75.6 $\pm$ 14.8 & 67.8 $\pm$ 18.8 & 42.6 $\pm$ 73 \\
\bottomrule
\end{tabular}
\end{table*}
\begin{table*}
\caption{\textbf{L* extraction} on the \textbf{Tomita} Languages with different RNN architectures on the \textbf{noisy dataset}. We highlight in bold the models that have obtained perfect performances for T1-T4. Since no model can reach perfect results on T5 and T6, for these languages we highlight in bold the model reaching the best RNN accuracy on the dev set. We omit the results for T7, since all the RNNs and the corresponding DFA achieve perfect performance on this language.} \label{tbl:L_star_Tomita_noisy}
\begin{tabular}{cccccccccc}
\toprule
\multirow{2}{*}{Model} & \multicolumn{4}{c}{T1 ($|Q| = 2$)} & \multicolumn{4}{c}{T2 ($|Q| = 3$)} \\
\cmidrule(lr){2-5} \cmidrule(lr){6-9}
& Train & Dev & Test & $|\hat{Q}|$ & Train & Dev & Test & $|\hat{Q}|$ \\
\midrule
        LSTM(1,30) & 99.5 $\pm$ 0 & 100 $\pm$ 0 & --- & --- & \textbf{99.5 $\pm$ 0.4} & \textbf{100 $\pm$ 0} & --- & --- \\
        DFA extracted & 99.9 $\pm$ 0.1 & 100 $\pm$ 0 & 100 $\pm$ 0 & 124.2 $\pm$ 273.2 & \textbf{100 $\pm$ 0} & \textbf{100 $\pm$ 0} & \textbf{100 $\pm$ 0} & \textbf{3 $\pm$ 0} \\
        \midrule
        LSTM(1,100) & \textbf{100 $\pm$ 0} & \textbf{100 $\pm$ 0} & --- & --- & \textbf{99.8 $\pm$ 0.3} & \textbf{100 $\pm$ 0} & --- & --- \\
        DFA extracted & \textbf{100 $\pm$ 0} & \textbf{100 $\pm$ 0} & \textbf{100 $\pm$ 0} & \textbf{2 $\pm$ 0} & \textbf{100 $\pm$ 0} & \textbf{100 $\pm$ 0} & \textbf{100 $\pm$ 0} & \textbf{3 $\pm$ 0} \\
        \midrule
        LSTM(2,100) & 100 $\pm$ 0 & 100 $\pm$ 0 & --- & --- & 100 $\pm$ 0 & 100 $\pm$ 0 & --- & --- \\
        DFA extracted & 99.6 $\pm$ 0.4 & 100 $\pm$ 0 & 100 $\pm$ 0 & 5.2 $\pm$ 4.6 & 98.7 $\pm$ 0 & 100 $\pm$ 0 & 100 $\pm$ 0 & 2 $\pm$ 0 \\
        \midrule
        GRU(1,30) & \textbf{99.5 $\pm$ 0.2} & \textbf{100 $\pm$ 0} & --- & --- & 99.6 $\pm$ 0.2 & 100 $\pm$ 0 & --- & --- \\
        DFA extracted & \textbf{100 $\pm$ 0} & \textbf{100 $\pm$ 0} & \textbf{100 $\pm$ 0} & \textbf{2 $\pm$ 0} & 100 $\pm$ 0 & 100 $\pm$ 0 & 100 $\pm$ 0 & 4 $\pm$ 2.2 \\
        \midrule
        GRU(1,100) & \textbf{99.8 $\pm$ 0.3} & \textbf{100 $\pm$ 0} & --- & --- & 100 $\pm$ 0 & 100 $\pm$ 0 & --- & --- \\
        DFA extracted & \textbf{100 $\pm$ 0} & \textbf{100 $\pm$ 0} & \textbf{100 $\pm$ 0} & \textbf{2 $\pm$ 0} & 99.6 $\pm$ 0.4 & 100 $\pm$ 0 & 100 $\pm$ 0 & 48.8 $\pm$ 78.6 \\
        \midrule
        GRU(2,100) & \textbf{100 $\pm$ 0} & \textbf{100 $\pm$ 0} & --- & --- & 100 $\pm$ 0 & 100 $\pm$ 0 & --- & --- \\
        DFA extracted & \textbf{100 $\pm$ 0} & \textbf{100 $\pm$ 0} & \textbf{100 $\pm$ 0} & \textbf{2 $\pm$ 0} & 98.9 $\pm$ 0.5 & 100 $\pm$ 0 & 100 $\pm$ 0 & 2.2 $\pm$ 0.4 \\
        \bottomrule
\end{tabular}
\centering
\setlength{\tabcolsep}{4pt}
\begin{tabular}{cccccccccc}
\toprule
\multirow{2}{*}{Model} & \multicolumn{4}{c}{T3 ($|Q| = 5$)} & \multicolumn{4}{c}{T4 ($|Q| = 4$)} \\
\cmidrule(lr){2-5} \cmidrule(lr){6-9}
& Train & Dev & Test & $|\hat{Q}|$ & Train & Dev & Test & $|\hat{Q}|$ \\
        \midrule
        LSTM(1,30) & \textbf{99.2 $\pm$ 0.2} & \textbf{100 $\pm$ 0} & --- & --- & \textbf{99.2 $\pm$ 0} & \textbf{100 $\pm$ 0} & --- & --- \\
        DFA extracted  & \textbf{100 $\pm$ 0} & \textbf{100 $\pm$ 0} & \textbf{100 $\pm$ 0} & \textbf{7.4 $\pm$ 3.3} & \textbf{100 $\pm$ 0} & \textbf{100 $\pm$ 0} & \textbf{100 $\pm$ 0} & \textbf{4 $\pm$ 0} \\
        \midrule
        LSTM(1,100) & 99.1 $\pm$ 0.9 & 100 $\pm$ 0 & --- & --- & \textbf{99.2 $\pm$ 0.3} & \textbf{100 $\pm$ 0} & --- & --- \\
        DFA extracted & 99.7 $\pm$ 0.2 & 100 $\pm$ 0 & 100 $\pm$ 0 & 161.8 $\pm$ 340 & \textbf{100 $\pm$ 0} & \textbf{100 $\pm$ 0} & \textbf{100 $\pm$ 0} & \textbf{4 $\pm$ 0} \\
        \midrule
        LSTM(2,100) & 99.2 $\pm$ 0.8 & 99 $\pm$ 2.2 & --- & --- & 99.4 $\pm$ 0 & 100 $\pm$ 0 & --- & --- \\
        DFA extracted & 99.5 $\pm$ 0.6 & 100 $\pm$ 0 & 100 $\pm$ 0 & 198 $\pm$ 254.8 & 99.9 $\pm$ 0 & 100 $\pm$ 0 & 100 $\pm$ 0 & 11.4 $\pm$ 5.7 \\
        \midrule
        GRU(1,30) & 99.1 $\pm$ 0.6 & 100 $\pm$ 0 & --- & --- & 99.2 $\pm$ 0 & 99.5 $\pm$ 1.1 & --- & --- \\
        DFA extracted & 100 $\pm$ 0.1 & 100 $\pm$ 0 & 100 $\pm$ 0 & 209.6 $\pm$ 282.1 & 100 $\pm$ 0 & 100 $\pm$ 0 & 100 $\pm$ 0 & 4 $\pm$ 0 \\
        \midrule
        GRU(1,100) & 99.5 $\pm$ 0.3 & 100 $\pm$ 0 & --- & --- & 99.4 $\pm$ 0.2 & 98 $\pm$ 2.1 & --- & --- \\
        DFA extracted & 99.6 $\pm$ 0.3 & 100 $\pm$ 0 & 100 $\pm$ 0 & 199.6 $\pm$ 258.1 & 99.9 $\pm$ 0 & 100 $\pm$ 0 & 99 $\pm$ 2.2 & 191.4 $\pm$ 297.5 \\
        \midrule
        GRU(2,100) & 99.8 $\pm$ 0.2 & 100 $\pm$ 0 & --- & --- & 99.7 $\pm$ 0.2 & 100 $\pm$ 0 & --- & --- \\
        DFA extracted & 99.7 $\pm$ 0.2 & 100 $\pm$ 0 & 100 $\pm$ 0 & 415.2 $\pm$ 313.9 & 100 $\pm$ 0 & 100 $\pm$ 0 & 100 $\pm$ 0 & 132.8 $\pm$ 288 \\
        \bottomrule
    \end{tabular}
\centering
\setlength{\tabcolsep}{2pt} 

    \begin{tabular}{cccccccccc}
        \toprule
        \multirow{2}{*}{Model} & \multicolumn{4}{c}{T5 ($|Q| = 4$)} & \multicolumn{4}{c}{T6 ($|Q| = 3$)} \\
        \cmidrule(lr){2-5} \cmidrule(lr){6-9}
        & Train & Dev & Test & $|\hat{Q}|$ & Train & Dev & Test & $|\hat{Q}|$ \\
        \midrule
        LSTM(1,30) & 76 $\pm$ 0 & 50 $\pm$ 0 & --- & --- & 87.3 $\pm$ 6.9 & 63.2 $\pm$ 20.7 & --- & --- \\
        DFA extracted & 53.7 $\pm$ 0 & 50 $\pm$ 0 & 50 $\pm$ 0 & 2 $\pm$ 0 & 83.7 $\pm$ 9.4 & 64.2 $\pm$ 20.7 & 59.5 $\pm$ 23.1 & 456.8 $\pm$ 255 \\
        \midrule
        LSTM(1,100) & 79 $\pm$ 0.7 & 49.8 $\pm$ 1.1 & --- & --- & \textbf{99.4 $\pm$ 0.1} & \textbf{99.4 $\pm$ 0.7} & --- & --- \\
        DFA extracted & 55.3 $\pm$ 2 & 49.2 $\pm$ 1.4 & 50 $\pm$ 0.7 & 300 $\pm$ 403.6 & \textbf{99.9 $\pm$ 0.1} & \textbf{100 $\pm$ 0} & \textbf{99.9 $\pm$ 0.1} & \textbf{107 $\pm$ 170.6} \\
        \midrule
        LSTM(2,100) & 84.7 $\pm$ 8.2 & 59.3 $\pm$ 22.7 & --- & --- & 96.9 $\pm$ 1.8 & 74.3 $\pm$ 14.8 & --- & --- \\
        DFA extracted & 63.3 $\pm$ 20.6 & 59.8 $\pm$ 22.5 & 61.1 $\pm$ 21.7 & 351.4 $\pm$ 328.4 & 97 $\pm$ 3.4 & 80.9 $\pm$ 12.4 & 75 $\pm$ 16.2 & 479.4 $\pm$ 321 \\
        \midrule
        GRU(1,30) & 78.9 $\pm$ 0.5 & 49.3 $\pm$ 2.5 & --- & --- & 87.3 $\pm$ 1.5 & 60.7 $\pm$ 2.1 & --- & --- \\
        DFA extracted & 54.7 $\pm$ 0.5 & 50.8 $\pm$ 2.5 & 48.2 $\pm$ 1.8 & 351 $\pm$ 344 & 84.8 $\pm$ 2.9 & 57.7 $\pm$ 3.7 & 52.2 $\pm$ 5.2 & 477.2 $\pm$ 170 \\
        \midrule
        GRU(1,100) & 92.9 $\pm$ 8.3 & 75.9 $\pm$ 26.4 & --- & --- & 99.5 $\pm$ 0.1 & 99.2 $\pm$ 1.3 & --- & --- \\
        DFA extracted & 84.8 $\pm$ 21.2 & 79.3 $\pm$ 28.3 & 81 $\pm$ 26 & 298.6 $\pm$ 408.8 & 99.9 $\pm$ 0.1 & 99.9 $\pm$ 0.1 & 99.9 $\pm$ 0.1 & 143.4 $\pm$ 234.2 \\
        \midrule
        GRU(2,100) & \textbf{99.7 $\pm$ 0.1} & \textbf{99.6 $\pm$ 0.4} & --- & --- & 96.1 $\pm$ 2.2 & 58.8 $\pm$ 4 & --- & --- \\
        DFA extracted  & \textbf{100 $\pm$ 0} & \textbf{100 $\pm$ 0} & \textbf{100 $\pm$ 0} & \textbf{4 $\pm$ 0} & 94.9 $\pm$ 1.1 & 67 $\pm$ 3.6 & 61 $\pm$ 4.4 & 288.8 $\pm$ 240.5 \\
        \bottomrule
    \end{tabular}
\end{table*}

 \begin{table*}[]
 \caption{\textbf{L* extraction} on the \textbf{Tomita 7} Language with different RNN architectures on the \textbf{noisy dataset}.}
     \label{tbl:T7_L*_noisy}
     \centering
\setlength{\tabcolsep}{3pt}
\begin{tabular}{ccccc} 
\toprule 
\multirow{2}{*}{Model} & \multicolumn{4}{c}{T7 ($|Q| = 5$)} \\ 
\cmidrule(lr){2-5}
& Train & Dev & Test & $|\hat{Q}|$ \\ 
\midrule 
LSTM(1,30) & 99.4 $\pm$ 0.1 & 99.0 $\pm$ 2.2 & --- & --- \\ 
DFA extracted &  100 $\pm$ 0 & 100 $\pm$ 0 & 100 $\pm$ 0 & 5 $\pm$ 0 \\ 
\midrule 
LSTM(1,100) & 99.4 $\pm$ 0.1 & 95.0 $\pm$ 5.0 & --- & --- \\ 
DFA extracted & 100 $\pm$ 0 & 100 $\pm$ 0 & 100 $\pm$ 0 & 5 $\pm$ 0 \\ 
\midrule 
LSTM(2,100) & 99.6 $\pm$ 0.08 & 99.0 $\pm$ 2.2 & --- & --- \\ 
DFA extracted & 100 $\pm$ 0 & 100 $\pm$ 0 & 100 $\pm$ 0 & 5 $\pm$ 0 \\ 
\midrule 
GRU(1,30) & 99.3 $\pm$ 0.1 & 100 $\pm$ 0 & --- & --- \\ 
DFA extracted & 100 $\pm$ 0 & 100 $\pm$ 0 & 100 $\pm$ 0 & 68.4 $\pm$ 141.7 \\ 
\midrule 
GRU(1,100) & \textbf{99.5 $\pm$ 0} & \textbf{100 $\pm$ 0} & --- & --- \\ 
DFA extracted & \textbf{100 $\pm$ 0} & \textbf{100 $\pm$ 0} & \textbf{100 $\pm$ 0} & \textbf{5 $\pm$ 0} \\ 
\midrule 
GRU(2,100) & \textbf{99.6} $\pm$ 0.2 & \textbf{100 $\pm$ 0} & --- & --- \\ 
DFA extracted & \textbf{100 $\pm$ 0} & \textbf{100 $\pm$ 0} & \textbf{100 $\pm$ 0} & \textbf{5 $\pm$ 0} \\ 
\bottomrule 
\end{tabular}
\end{table*}

\begin{table*}[t!]
\centering
\caption{\textbf{DeepDFA} experiments on the \textbf{Tomita} Languages with different sizes of the state space. We highlight in bold the models that have obtained perfect performances.} \label{tbl:deepDFA_Tomita}
\begin{tabular}{cccccccccc} 
\toprule 
\multirow{2}{*}{Model} & \multicolumn{4}{c}{T1 ($|Q| = 2$)} & \multicolumn{4}{c}{T2 ($|Q| = 3$)} \\ 
\cmidrule(lr){2-5} \cmidrule(lr){6-9}
& Train & Dev & Test & $|\hat{Q}|$ & Train & Dev & Test & $|\hat{Q}|$ \\ 
\midrule 
DeepDFA(10) & \textbf{100 $\pm$ 0} & \textbf{100 $\pm$ 0} & --- & --- & \textbf{100 $\pm$ 0} & \textbf{100 $\pm$ 0} & --- & --- \\ 
DFA extracted & \textbf{100 $\pm$ 0} & \textbf{100 $\pm$ 0} & \textbf{100 $\pm$ 0} & \textbf{2 $\pm$ 0} & \textbf{100 $\pm$ 0} & \textbf{100 $\pm$ 0} & \textbf{100 $\pm$ 0} & \textbf{3 $\pm$ 0} \\ 
\midrule 
DeepDFA(30) & \textbf{100 $\pm$ 0} & \textbf{100 $\pm$ 0} & --- & --- & \textbf{100 $\pm$ 0} & \textbf{100 $\pm$ 0} & --- & --- \\ 
DFA extracted & \textbf{100 $\pm$ 0} & \textbf{100 $\pm$ 0} & \textbf{100 $\pm$ 0} & \textbf{2 $\pm$ 0} & \textbf{100 $\pm$ 0} & \textbf{100 $\pm$ 0} & \textbf{100 $\pm$ 0} & \textbf{3 $\pm$ 0} \\ 
\midrule 
DeepDFA(100) & \textbf{100 $\pm$ 0} & \textbf{100 $\pm$ 0} & --- & --- & \textbf{100 $\pm$ 0} & \textbf{100 $\pm$ 0} & --- & --- \\ 
DFA extracted & \textbf{100 $\pm$ 0} & \textbf{100 $\pm$ 0} & \textbf{100 $\pm$ 0} & \textbf{2 $\pm$ 0} & \textbf{100 $\pm$ 0} & \textbf{100 $\pm$ 0} & \textbf{100 $\pm$ 0} & \textbf{3 $\pm$ 0} \\ 
\bottomrule 
\end{tabular}

\vspace{0.2cm}

\begin{tabular}{cccccccccc} 
\toprule 
\multirow{2}{*}{Model} & \multicolumn{4}{c}{T3 ($|Q| = 5$)} & \multicolumn{4}{c}{T4 ($|Q| = 4$)} \\ 
\cmidrule(lr){2-5} \cmidrule(lr){6-9}
& Train & Dev & Test & $|\hat{Q}|$ & Train & Dev & Test & $|\hat{Q}|$ \\ 
\midrule 
DeepDFA(10) & \textbf{100 $\pm$ 0} & \textbf{100 $\pm$ 0} & --- & --- & \textbf{100 $\pm$ 0} & \textbf{100 $\pm$ 0} & --- & --- \\ 
DFA extracted & \textbf{100 $\pm$ 0} & \textbf{100 $\pm$ 0} & \textbf{100 $\pm$ 0} & \textbf{5 $\pm$ 0} & \textbf{100 $\pm$ 0} & \textbf{100 $\pm$ 0} & \textbf{100 $\pm$ 0} & \textbf{4 $\pm$ 0} \\ 
\midrule 
DeepDFA(30) & 100 $\pm$ 0 & 100 $\pm$ 0 & --- & --- & \textbf{100 $\pm$ 0} & \textbf{100 $\pm$ 0} & --- & --- \\ 
DFA extracted & 98.8 $\pm$ 2.6 & 100 $\pm$ 0 & 100 $\pm$ 0 & 4.8 $\pm$ 0.4 & \textbf{100 $\pm$ 0} & \textbf{100 $\pm$ 0} & \textbf{100 $\pm$ 0} & \textbf{4 $\pm$ 0} \\ 
\midrule 
DeepDFA(100) & 99.3 $\pm$ 1.3 & 100 $\pm$ 0 & --- & --- & \textbf{100 $\pm$ 0} & \textbf{100 $\pm$ 0} & --- & --- \\ 
DFA extracted & 90.4 $\pm$ 13 & 100 $\pm$ 0 & 100 $\pm$ 0 & 4.2 $\pm$ 1.1 & \textbf{100 $\pm$ 0} & \textbf{100 $\pm$ 0} & \textbf{100 $\pm$ 0} & \textbf{4 $\pm$ 0} \\ 
\bottomrule 
\end{tabular}

\vspace{0.2cm}

\begin{tabular}{cccccccccc} 
\toprule 
\multirow{2}{*}{Model} & \multicolumn{4}{c}{T5 ($|Q| = 4$)} & \multicolumn{4}{c}{T6 ($|Q| = 3$)} \\ 
\cmidrule(lr){2-5} \cmidrule(lr){6-9}
& Train & Dev & Test & $|\hat{Q}|$ & Train & Dev & Test & $|\hat{Q}|$ \\ 
\midrule 
DeepDFA(10) & \textbf{100 $\pm$ 0} & \textbf{100 $\pm$ 0} & --- & --- & \textbf{100 $\pm$ 0} & \textbf{100 $\pm$ 0} & --- & --- \\ 
DFA extracted & \textbf{100 $\pm$ 0} & \textbf{100 $\pm$ 0} & \textbf{100 $\pm$ 0} & \textbf{4 $\pm$ 0} & \textbf{100 $\pm$ 0} & \textbf{100 $\pm$ 0} & \textbf{100 $\pm$ 0} & \textbf{3 $\pm$ 0} \\ 
\midrule 
DeepDFA(30) & \textbf{100 $\pm$ 0} & \textbf{100 $\pm$ 0} & --- & --- & \textbf{100 $\pm$ 0} & \textbf{100 $\pm$ 0} & --- & --- \\ 
DFA extracted & \textbf{100 $\pm$ 0} & \textbf{100 $\pm$ 0} & \textbf{100 $\pm$ 0} & \textbf{4 $\pm$ 0} & \textbf{100 $\pm$ 0} & \textbf{100 $\pm$ 0} & \textbf{100 $\pm$ 0} & \textbf{3 $\pm$ 0} \\ 
\midrule 
DeepDFA(100) & \textbf{100 $\pm$ 0} & \textbf{100 $\pm$ 0} & --- & --- & \textbf{100 $\pm$ 0} &  \textbf{100 $\pm$ 0} & --- & --- \\ 
DFA extracted & \textbf{100 $\pm$ 0} & \textbf{100 $\pm$ 0} & \textbf{100 $\pm$ 0} & \textbf{4 $\pm$ 0} & \textbf{100 $\pm$ 0} & \textbf{100 $\pm$ 0} & \textbf{100 $\pm$ 0} & \textbf{3 $\pm$ 0} \\ 
\bottomrule 
\end{tabular}

\vspace{0.2cm}

\begin{tabular}{ccccc} 
\toprule 
\multirow{2}{*}{Model} & \multicolumn{4}{c}{T7 ($|Q| = 5$)} \\ 
\cmidrule(lr){2-5}
& Train & Dev & Test & $|\hat{Q}|$ \\ 
\midrule 
DeepDFA(10) & \textbf{100 $\pm$ 0} & \textbf{100 $\pm$ 0} & --- & --- \\ 
DFA extracted & \textbf{100 $\pm$ 0} & \textbf{100 $\pm$ 0} & \textbf{100 $\pm$ 0} & \textbf{5 $\pm$ 0} \\ 
\midrule 
DeepDFA(30) & 100 $\pm$ 0 & 100 $\pm$ 0 & --- & --- \\ 
DFA extracted & 96.4 $\pm$ 7.9 & 100 $\pm$ 0 & 100 $\pm$ 0 & 5.2 $\pm$ 0.4 \\ 
\midrule 
DeepDFA(100) & \textbf{100 $\pm$ 0} & \textbf{100 $\pm$ 0} & --- & --- \\ 
DFA extracted & \textbf{100 $\pm$ 0} & \textbf{100 $\pm$ 0} & \textbf{100 $\pm$ 0} & \textbf{5 $\pm$ 0} \\ 
\bottomrule 
\end{tabular}
\end{table*}


\begin{table*}[t!]
\centering
\caption{\textbf{DeepDFA} experiments on the \textbf{noisy dataset} constructed for the \textbf{Tomita} Languages. We test the model with different sizes of the state space. We highlight in bold the models that have obtained perfect accuracy on the dev and test set with both the RNN and the DFA and with the number of states of the predicted DFA equal to the target one.}
\label{tbl:deepdfa_Tomita_noisy}
\begin{tabular}{cccccccccc}
\toprule
\multirow{2}{*}{Model} & \multicolumn{4}{c}{T1 ($|Q| = 2$)} & \multicolumn{4}{c}{T2 ($|Q| = 3$)} \\
\cmidrule(lr){2-5} \cmidrule(lr){6-9}
 & Train & Dev & Test & $|\hat{Q}|$ & Train & Dev & Test & $|\hat{Q}|$ \\
\midrule
DeepDFA(10) & 99.2 $\pm$ 0.1 & 100 $\pm$ 0 & --- & --- & \textbf{99 $\pm$ 0} & \textbf{100 $\pm$ 0} & --- & --- \\
DFA extracted & 98.3 $\pm$ 1.2 & 99 $\pm$ 2.2 & 99 $\pm$ 2.2 & 5.2 $\pm$ 1.7 & \textbf{99 $\pm$ 0} & \textbf{100 $\pm$ 0 }& \textbf{100 $\pm$ 0 }& \textbf{3 $\pm$ 0} \\
\midrule
DeepDFA(30) & \textbf{98.9 $\pm$ 0.2} & \textbf{100 $\pm$ 0} & --- & --- & \textbf{98.7 $\pm$ 0.9} & \textbf{100 $\pm$ 0} & --- & --- \\
DFA extracted & \textbf{99 $\pm$ 00} & \textbf{100 $\pm$ 0} & \textbf{100 $\pm$ 0} & \textbf{2 $\pm$ 0} & \textbf{100 $\pm$ 0} & \textbf{100 $\pm$ 0} & \textbf{100 $\pm$ 0} & \textbf{3 $\pm$ 0} \\
\midrule
DeepDFA(100) & 99 $\pm$ 0 & 100 $\pm$ 0 & --- & --- & 97.1 $\pm$ 3.2 & 96 $\pm$ 5.4 & --- & --- \\
DFA extracted & 93.9 $\pm$ 11.4 & 94 $\pm$ 13.4 & 97 $\pm$ 6.7 & 2.8 $\pm$ 1.7 & 83.3 $\pm$ 8 & 86 $\pm$ 8.2 & 85 $\pm$ 9.3 & 10 $\pm$ 1.8 \\
\midrule
\end{tabular}
\vspace{0.2cm}

\begin{tabular}{cccccccccc}
\toprule
\multirow{2}{*}{Model} & \multicolumn{4}{c}{T3 ($|Q| = 5$)} & \multicolumn{4}{c}{T4 ($|Q| = 4$)} \\
\cmidrule(lr){2-5} \cmidrule(lr){6-9}
 & Train & Dev & Test & $|\hat{Q}|$ & Train & Dev & Test & $|\hat{Q}|$ \\
\midrule
DeepDFA(10) & \textbf{99 $\pm$ 0} & \textbf{100 $\pm$ 0} & --- & --- & \textbf{98.9 $\pm$ 0} & \textbf{100 $\pm$ 0} & --- & --- \\
DFA extracted & \textbf{99 $\pm$ 0} & \textbf{100 $\pm$ 0} & \textbf{100 $\pm$ 0} & \textbf{5 $\pm$ 0} & \textbf{98.9 $\pm$ 0} & \textbf{100 $\pm$ 0} & \textbf{100 $\pm$ 0} & \textbf{4 $\pm$ 0} \\
\midrule
DeepDFA(30) & \textbf{99 $\pm$ 0} & \textbf{100 $\pm$ 0} & --- & --- & \textbf{98.9 $\pm$ 0} & \textbf{100 $\pm$ 0} & --- & --- \\
DFA extracted & \textbf{99 $\pm$ 0} & \textbf{100 $\pm$ 0} & \textbf{100 $\pm$ 0} & \textbf{5 $\pm$ 0} & \textbf{98.9 $\pm$ 0} & \textbf{100 $\pm$ 0} & \textbf{100 $\pm$ 0} & \textbf{4 $\pm$ 0} \\
\midrule
DeepDFA(100) & 97.2 $\pm$ 2.8 & 100 $\pm$ 0 & --- & --- & \textbf{98.9 $\pm$ 0} & \textbf{100 $\pm$ 0} & --- & --- \\
DFA extracted & 95.1 $\pm$ 5 & 95 $\pm$ 11.1 & 96 $\pm$ 8.9 & 8 $\pm$ 3 & \textbf{98.9 $\pm$ 0} & \textbf{100 $\pm$ 0} & \textbf{100 $\pm$ 0} & \textbf{4 $\pm$ 0} \\
\midrule
\end{tabular}
\vspace{0.2cm}

\begin{tabular}{cccccccccc}
\toprule
\multirow{2}{*}{Model} & \multicolumn{4}{c}{T5 ($|Q| = 4$)} & \multicolumn{4}{c}{T6 ($|Q| = 3$)} \\
\cmidrule(lr){2-5} \cmidrule(lr){6-9}
 & Train & Dev & Test & $|\hat{Q}|$ & Train & Dev & Test & $|\hat{Q}|$ \\
\midrule
DeepDFA(10) & \textbf{99 $\pm$ 0} & \textbf{100 $\pm$ 0} & --- & --- & \textbf{99 $\pm$ 0} & \textbf{100 $\pm$ 0} & --- & --- \\
DFA extracted & \textbf{99 $\pm$ 0} & \textbf{100 $\pm$ 0} & \textbf{100 $\pm$ 0} & \textbf{3 $\pm$ 0} & \textbf{99 $\pm$ 0} & \textbf{100 $\pm$ 0} & \textbf{100 $\pm$ 0} & \textbf{3 $\pm$ 0} \\
\midrule
DeepDFA(30) & \textbf{99 $\pm$ 0} & \textbf{100 $\pm$ 0} & --- & --- & \textbf{99 $\pm$ 0} & \textbf{100 $\pm$ 0} & --- & --- \\
DFA extracted & \textbf{99 $\pm$ 0} & \textbf{100 $\pm$ 0} & \textbf{100 $\pm$ 0} & \textbf{3 $\pm$ 0} & \textbf{99 $\pm$ 0} & \textbf{100 $\pm$ 0} & \textbf{100 $\pm$ 0} & \textbf{3 $\pm$ 0} \\
\midrule
DeepDFA(100) & 99 $\pm$ 0 & 99.9 $\pm$ 0.1 & --- & --- & \textbf{99 $\pm$ 0} & \textbf{100 $\pm$ 0} & --- & --- \\
DFA extracted & 99 $\pm$ 0 & 100 $\pm$ 0 & 100 $\pm$ 0 & 3 $\pm$ 0 & \textbf{99 $\pm$ 0} & \textbf{100 $\pm$ 0} & \textbf{100 $\pm$ 0} & \textbf{3 $\pm$ 0} \\
\midrule
\end{tabular}
\vspace{0.2cm}

\begin{tabular}{cccccccccc}
\toprule
\multirow{2}{*}{Model} & \multicolumn{4}{c}{T7 ($|Q| = 5$)} \\
\cmidrule(lr){2-5}
 & Train & Dev & Test & $|\hat{Q}|$ \\
\midrule
DeepDFA(10) & \textbf{99 $\pm$ 0} & \textbf{100 $\pm$ 0} & --- & --- \\
DFA extracted & \textbf{100 $\pm$ 0} & \textbf{100 $\pm$ 0} & \textbf{100 $\pm$ 0} & \textbf{5 $\pm$ 0} \\
\midrule
DeepDFA(30) & 99 $\pm$ 0 & 100 $\pm$ 0 & --- & --- \\
DFA extracted & 98.8 $\pm$ 0.4 & 100 $\pm$ 0 & 100 $\pm$ 0 & 6.2 $\pm$ 2.6 \\
\midrule
DeepDFA(100) & \textbf{99 $\pm$ 0} & \textbf{100 $\pm$ 0} & --- & --- \\
DFA extracted & \textbf{99 $\pm$ 0} & \textbf{100 $\pm$ 0} & \textbf{100 $\pm$ 0} & \textbf{5 $\pm$ 0} \\
\bottomrule
\end{tabular}
\end{table*}


\begin{table*}[t!]
\centering
\caption{\textbf{DFA Generator} experiments on the \textbf{Tomita} Languages with different sizes of the state space. We highlight in bold the models that have obtained perfect performances.} \label{tbl:deepDFA_Tomita}
\begin{tabular}{cccccccccc} 
\toprule 
\multirow{2}{*}{Model} & \multicolumn{4}{c}{T1 ($|Q| = 2$)} & \multicolumn{4}{c}{T2 ($|Q| = 3$)} \\ 
\cmidrule(lr){2-5} \cmidrule(lr){6-9}
& Train & Dev & Test & $|\hat{Q}|$ & Train & Dev & Test & $|\hat{Q}|$ \\ 
\midrule 
DFAGen(10) & 100 $\pm$ 0 & 98.5 $\pm$ 1.8 & --- & --- & 98.7 $\pm$ 0 & 100 $\pm$ 0 & --- & --- \\ 
DFA extracted & 97 $\pm$ 0.8 & 100 $\pm$ 0 & 100 $\pm$ 0 & 1.8 $\pm$ 1.1 & 98.7 $\pm$ 0 & 100 $\pm$ 0 & 100 $\pm$ 0 & 1 $\pm$ 0 \\ 
\midrule 
DFAGen(30) & 98.8 $\pm$ 1.5 & 100 $\pm$ 0 & --- & --- & 98.7 $\pm$ 0 & 100 $\pm$ 0 & --- & --- \\ 
DFA extracted &  96.9 $\pm$ 0.8 & 100 $\pm$ 0 & 100 $\pm$ 0 & 1.4 $\pm$ 0.8 & 98.7 $\pm$ 0 & 100 $\pm$ 0 & 100 $\pm$ 0 & 1 $\pm$ 0 \\ 
\midrule 
DeepDFA(100) & \textbf{Out Of time} & --- & --- & --- & --- & --- & --- & --- \\ 

\bottomrule 
\end{tabular}

\vspace{0.2cm}

\begin{tabular}{cccccccccc} 
\toprule 
\multirow{2}{*}{Model} & \multicolumn{4}{c}{T3 ($|Q| = 5$)} & \multicolumn{4}{c}{T4 ($|Q| = 4$)} \\ 
\cmidrule(lr){2-5} \cmidrule(lr){6-9}
& Train & Dev & Test & $|\hat{Q}|$ & Train & Dev & Test & $|\hat{Q}|$ \\ 
\midrule 
DFAGen(10) & 99.9 $\pm$ 0 & 100 $\pm$ 0 & --- & --- & 100 $\pm$ 0 & 97.5 $\pm$ 0 & --- & --- \\ 
DFA extracted & 89.1 $\pm$ 7.8 & 90 $\pm$ 17.3 & 91 $\pm$ 15.1 & 5 $\pm$ 1.5 & 87.7 $\pm$ 18.3 & 82 $\pm$ 28.4 & 85 $\pm$ 33.5 & 4.8 $\pm$ 0.4 \\ 
\midrule 
DFAGen(30) & 99.8 $\pm$ 0.2 & 99 $\pm$ 2.2 & --- & --- & 99.9 $\pm$ 0 & 97.5 $\pm$ 0 & --- & --- \\ 
DFA extracted & 96 $\pm$ 8.8 & 97 $\pm$ 6.7 & 87 $\pm$ 29 & 5.2 $\pm$ 0.4 & 84.4 $\pm$ 21.3 & 83 $\pm$ 26.3 & 82 $\pm$ 40.2 & 4.8 $\pm$ 0.4 \\ 
\midrule 
DFAGen(100) & \textbf{Out Of Time} & --- & --- & --- & --- & --- & --- \\ 
\bottomrule 
\end{tabular}

\vspace{0.2cm}

\begin{tabular}{cccccccccc} 
\toprule 
\multirow{2}{*}{Model} & \multicolumn{4}{c}{T5 ($|Q| = 4$)} & \multicolumn{4}{c}{T6 ($|Q| = 3$)} \\ 
\cmidrule(lr){2-5} \cmidrule(lr){6-9}
& Train & Dev & Test & $|\hat{Q}|$ & Train & Dev & Test & $|\hat{Q}|$ \\ 
\midrule 
DFAGen(10) & 96.2 $\pm$ 0 & 100 $\pm$ 0 & --- & --- & \textbf{100 $\pm$ 0} & \textbf{100 $\pm$ 0} & --- & --- \\ 
DFA extracted & 96.2 $\pm$ 0 & 100 $\pm$ 0 & 100 $\pm$ 0 & 2.6 $\pm$ 0.8 & \textbf{100 $\pm$ 0} & \textbf{100 $\pm$ 0} & \textbf{100 $\pm$ 0} & \textbf{3 $\pm$ 0} \\ 
\midrule 
DFAGen(30) & \textbf{Out Of Time} & --- & --- & --- & \textbf{100 $\pm$ 0} & \textbf{100 $\pm$ 0} & \textbf{100 $\pm$ 0}& --- & --- \\ 
DFA extracted & --- & --- &  --- & --- & \textbf{100 $\pm$ 0} & \textbf{100 $\pm$ 0} & \textbf{100 $\pm$ 0} & \textbf{3 $\pm$ 0} \\ 
\midrule 
DFAGen(100) & \textbf{Out Of Time} & --- & --- & --- & --- &  --- & --- & --- \\ 
\bottomrule 
\end{tabular}

\vspace{0.2cm}

\begin{tabular}{ccccc} 
\toprule 
\multirow{2}{*}{Model} & \multicolumn{4}{c}{T7 ($|Q| = 5$)} \\ 
\cmidrule(lr){2-5}
& Train & Dev & Test & $|\hat{Q}|$ \\ 
\midrule 
DFAGen(10) & 98 $\pm$ 2.3 & 100 $\pm$ 0 & --- & --- \\ 
DFA extracted & 83.1 $\pm$ 22.8 & 100 $\pm$ 0 & 100 $\pm$ 0 & 4.4 $\pm$ 0.8 \\ 
\midrule 
DFAGen(30) & \textbf{Out Of Time} & --- & --- & --- \\ 
\midrule 
DFAGen(100) & \textbf{Out Of Time} & --- & --- & --- \\ 

\bottomrule 
\end{tabular}
\end{table*}


\begin{table*}[t!]
\centering
\caption{\textbf{DFA Generator} experiments on the \textbf{noisy dataset} constructed for the \textbf{Tomita} Languages. We test the model with different sizes of the state space. We highlight in bold the models that have obtained perfect accuracy on the dev and test set with both the RNN and the DFA and with the number of states of the predicted DFA equal to the target one.}
\label{tbl:deepdfa_Tomita_noisy}
\begin{tabular}{cccccccccc} 
\toprule 
\multirow{2}{*}{Model} & \multicolumn{4}{c}{T1 ($|Q| = 2$)} & \multicolumn{4}{c}{T2 ($|Q| = 3$)} \\ 
\cmidrule(lr){2-5} \cmidrule(lr){6-9}
& Train & Dev & Test & $|\hat{Q}|$ & Train & Dev & Test & $|\hat{Q}|$ \\ 
\midrule 
DFAGen(10) & 98.4 $\pm$ 1.38 & 100 $\pm$ 0 & --- & --- & 97.7 $\pm$ 0 & 100 $\pm$ 0 & --- & --- \\ 
DFA extracted & 96.7 $\pm$ 1.4 & 100 $\pm$ 0 & 100 $\pm$ 0 & 2.2 $\pm$ 1.1 & 97.7 $\pm$ 0 & 100 $\pm$ 0 & 100 $\pm$ 0 & 1 $\pm$ 0 \\ 
\midrule 
DFAGen(30) & 97.1 $\pm$ 1.7 & 100 $\pm$ 0 & --- & --- & 98 $\pm$ 0.4 & 100 $\pm$ 0 & --- & --- \\ 
DFA extracted &  96.4 $\pm$ 1.0 & 100 $\pm$ 0 & 100 $\pm$ 0 & 1.8 $\pm$ 1.1 & 97.6 $\pm$ 0.4 & 100 $\pm$ 0 & 100 $\pm$ 0 & 2.4 $\pm$ 1.9 \\ 
\midrule 
DeepDFA(100) & \textbf{Out Of time} & --- & --- & --- & --- & --- & --- & --- \\ 

\bottomrule 
\end{tabular}

\vspace{0.2cm}

\begin{tabular}{cccccccccc} 
\toprule 
\multirow{2}{*}{Model} & \multicolumn{4}{c}{T3 ($|Q| = 5$)} & \multicolumn{4}{c}{T4 ($|Q| = 4$)} \\ 
\cmidrule(lr){2-5} \cmidrule(lr){6-9}
& Train & Dev & Test & $|\hat{Q}|$ & Train & Dev & Test & $|\hat{Q}|$ \\ 
\midrule 
DFAGen(10) & 98.9 $\pm$ 0 & 100 $\pm$ 0 & --- & --- & 100 $\pm$ 0 & 98.9 $\pm$ 0.1 & --- & --- \\ 
DFA extracted & 95.4 $\pm$ 4.1.8 & 100 $\pm$ 0 & 100 $\pm$ 0 & 4.8 $\pm$ 0.8 & 94 $\pm$ 11 & 95 $\pm$ 11.1 & 100 $\pm$ 0 & 4.8 $\pm$ 0.4 \\ 
\midrule 
DFAGen(30) & 98.9 $\pm$ 0 & 100 $\pm$ 0 & --- & --- & 98.9 $\pm$ 0 & 99 $\pm$ 1.3 & --- & --- \\ 
DFA extracted & 95.4 $\pm$ 3.2 & 100 $\pm$ 0 & 100 $\pm$ 0 & 4.6 $\pm$ 0.5 & 92.5 $\pm$ 6.9 & 95 $\pm$ 8.6 & 90 $\pm$ 22.3 & 5.8 $\pm$ 1.3 \\ 
\midrule 
DFAGen(100) & \textbf{Out Of Time} & --- & --- & --- & --- & --- & --- \\ 
\bottomrule 
\end{tabular}

\vspace{0.2cm}

\begin{tabular}{cccccccccc} 
\toprule 
\multirow{2}{*}{Model} & \multicolumn{4}{c}{T5 ($|Q| = 4$)} & \multicolumn{4}{c}{T6 ($|Q| = 3$)} \\ 
\cmidrule(lr){2-5} \cmidrule(lr){6-9}
& Train & Dev & Test & $|\hat{Q}|$ & Train & Dev & Test & $|\hat{Q}|$ \\ 
\midrule 
DFAGen(10) & 95.3 $\pm$ 0 & 99.9 $\pm$ 0.1 & --- & --- & 99 $\pm$ 0 & 100 $\pm$ 0 & --- & --- \\ 
DFA extracted & 95.3 $\pm$ 0.1 & 100 $\pm$ 0 & 100 $\pm$ 0 & 2.6 $\pm$ 0.8 & 99 $\pm$ 0 & 100 $\pm$ 0 & 100 $\pm$ 0 & 3 $\pm$ 0 \\ 
\midrule 
DFAGen(30) & \textbf{Out Of Time} & --- & --- & --- &  \textbf{99 $\pm$ 0} & \textbf{100 $\pm$ 0} & \textbf{100 $\pm$ 0}\\ 
DFA extracted & --- & --- &  --- & --- & \textbf{99 $\pm$ 0} & \textbf{100 $\pm$ 0} & \textbf{100 $\pm$ 0} & \textbf{3 $\pm$ 0} \\ 
\midrule 
DFAGen(100) & \textbf{Out Of Time} & --- & --- & --- & --- &  --- & --- & --- \\ 
\bottomrule 
\end{tabular}

\vspace{0.2cm}

\begin{tabular}{ccccc} 
\toprule 
\multirow{2}{*}{Model} & \multicolumn{4}{c}{T7 ($|Q| = 5$)} \\ 
\cmidrule(lr){2-5}
& Train & Dev & Test & $|\hat{Q}|$ \\ 
\midrule 
DFAGen(10) & 96.1 $\pm$ 2.7 & 100 $\pm$ 0 & --- & --- \\ 
DFA extracted & 74.1 $\pm$ 22.4 & 100 $\pm$ 0 & 100 $\pm$ 0 & 4.4 $\pm$ 1.4 \\ 
\midrule 
DFAGen(30) & \textbf{Out Of Time} & --- & --- & --- \\ 
\midrule 
DFAGen(100) & \textbf{Out Of Time} & --- & --- & --- \\ 

\bottomrule 
\end{tabular}
\end{table*}

 
\begin{table*}[t!]
    \centering
    \caption{\textbf{L* extraction}, \textbf{DeepDFA} and \textbf{DFA-generator} on the \textbf{random DFAs} benchmark, with different RNN architectures. We highlight in bold the model obtaining the best accuracy on the dev set for each approach.}
    \label{tbl:random_dfa}
    \centering
\setlength{\tabcolsep}{4pt}
    \begin{tabular}{cccccccccc}
        \toprule
        \multirow{2}{*}{Model} & \multicolumn{4}{c}{Random DFA $|Q| = 10$, $|P| = 2$} & \multicolumn{4}{c}{Random DFA $|Q| = 10$, $|P| = 3$} \\
        \cmidrule(lr){2-5} \cmidrule(lr){6-9}
        & Train & Dev & Test & $|\hat{Q}|$ & Train & Dev & Test & $|\hat{Q}|$ \\
        \midrule
        LSTM(1,100) & 100 $\pm$ 0 & 99.9 $\pm$ 0.2 & --- & --- & 100 $\pm$ 0 & 90.5 $\pm$ 7.7 & --- & --- \\
        DFA extracted & 96.2 $\pm$ 7.7 & 96.2 $\pm$ 7.5 & 96.2 $\pm$ 7.9 & 89.3 $\pm$ 221.3 & 96.1 $\pm$ 4.1 & 92.6 $\pm$ 7.3 & 91.6 $\pm$ 8.9 & 304 $\pm$ 233 \\
        \midrule
        LSTM(2,100) & 100 $\pm$ 0 & 99.9 $\pm$ 0.2 & --- & --- & \textbf{100 $\pm$ 0} & \textbf{95.4 $\pm$ 4.8} & --- & --- \\
        DFA extracted & 95.2 $\pm$ 8.6 & 94.9 $\pm$ 8.7 & 95.8 $\pm$ 7.6  & 120 $\pm$ 231.2 & \textbf{98.5 $\pm$ 2.4} & \textbf{96.5 $\pm$ 4.8} & \textbf{96.3 $\pm$ 5.6} & \textbf{282.1 $\pm$ 215.1} \\
        \midrule
        GRU(1,100) & 100 $\pm$ 0 & 99.9 $\pm$ 0.3 & --- & --- & 100$\pm$ 0 & 91.6 $\pm$ 8.5 & --- & --- \\
        DFA extracted & 99.9 $\pm$ 0.2 & 99.7 $\pm$ 0.6 & 99.8 $\pm$ 0.4 & 34.9 $\pm$ 116.1 & 97.9 $\pm$ 3.4 & 95 $\pm$ 7 & 94.7 $\pm$ 6.9 & 309.1 $\pm$ 178.7 \\
        \midrule
        GRU(2,100) & \textbf{100 $\pm$ 0} & \textbf{99.9 $\pm$ 0.1} & --- & --- & 100 $\pm$ 0 & 90.9 $\pm$ 11.8 & --- & --- \\
        DFA extracted & \textbf{96.4 $\pm$ 7} & \textbf{95.9 $\pm$ 7.6} & \textbf{96.8 $\pm$ 6.7} & \textbf{58.6 $\pm$ 176.2} & 96.4 $\pm$ 5.9  & 91.5 $\pm$ 11.6 & 91.9 $\pm$ 11.6 & 278.7 $\pm$ 221.6 \\
        \midrule
        
        \midrule
deepDFA(100) & \textbf{99.7} $\pm$ \textbf{1.1} & \textbf{99.6} $\pm$ \textbf{1.6} & --- & --- & \textbf{99.9} $\pm$ \textbf{0.0} & \textbf{99.9} $\pm$ \textbf{0.1} & --- & --- \\
DFA extracted & \textbf{97.8} $\pm$ \textbf{6.3} & \textbf{97.8} $\pm$ \textbf{6.5} & \textbf{97.7} $\pm$ \textbf{6.8} & \textbf{8.8} $\pm$ \textbf{2.9} & \textbf{99.5} $\pm$ \textbf{1.3} & \textbf{99.3} $\pm$ \textbf{2.0} & \textbf{99.4} $\pm$ \textbf{1.7} & \textbf{11.9} $\pm$ \textbf{4.6} \\
        \midrule

        \midrule
DFAGen(50) & \textbf{89.4} $\pm$ \textbf{20.1} & \textbf{89.4 }$\pm$ \textbf{20.1} & --- & --- & \textbf{o.o.t.} & --- & --- & --- \\
DFA extracted & \textbf{87.1} $\pm$ \textbf{19.3} & \textbf{86.8} $\pm$ \textbf{19.6} & \textbf{87 }$\pm$ \textbf{19.7} & \textbf{16 }$\pm$ \textbf{17.5} & --- & --- & --- & --- \\
        \midrule
DFAGen(100) & \textbf{o.o.t.} & ---  & --- & --- & ---  & ---  & --- & --- \\

        \bottomrule
    \end{tabular}
    \bigskip

        \centering
\setlength{\tabcolsep}{4pt}
    \begin{tabular}{cccccccccc}
        \toprule
        \multirow{2}{*}{Model} & \multicolumn{4}{c}{Random DFA $|Q| = 20$, $|P| = 2$} & \multicolumn{4}{c}{Random DFA $|Q| = 20$, $|P| = 3$} \\
        \cmidrule(lr){2-5} \cmidrule(lr){6-9}
        & Train & Dev & Test & $|\hat{Q}|$ & Train & Dev & Test & $|\hat{Q}|$ \\
        \midrule
        LSTM(1,100) & 100 $\pm$ 0 & 99.7 $\pm$ 0.4 & --- & --- &  98.9 $\pm$ 3.3 & 64 $\pm$ 3.6 & --- & --- \\
        DFA extracted & 99.9 $\pm$ 0 & 99.8 $\pm$ 0.2 & 99.8 $\pm$ 0.3 & 258.7 $\pm$ 321.3 & 72.1 $\pm$ 4.1 & 63.2 $\pm$ 3.4 & 64.9 $\pm$ 3.9 & 396.2 $\pm$ 213.5 \\
        \midrule
        LSTM(2,100) & 100 $\pm$ 0 & 99.6 $\pm$ 0.5 & --- & --- & \textbf{100 $\pm$ 0} & \textbf{69.3 $\pm$ 3.7} & --- & --- \\
        DFA extracted & 98.1 $\pm$ 8.7 & 97.8 $\pm$ 9 & 97.7 $\pm$ 9 & 315.7 $\pm$ 305.5 & \textbf{75.5 $\pm$ 4.4} & \textbf{65.3 $\pm$ 4.3} & \textbf{68.4 $\pm$ 4.7} & \textbf{405.3 $\pm$ 207.9} \\
        \midrule
        GRU(1,100) & 99.8 $\pm$ 0.8 & 99 $\pm$ 3.4 & --- & --- & 99.7 $\pm$ 1.6 & 65.7 $\pm$ 3.7 & --- & --- \\
        DFA extracted & 99.4 $\pm$ 2.6 & 98.8 $\pm$ 4.6 & 98.8 $\pm$ 4.6 & 157.9 $\pm$ 254.4 & 72.1 $\pm$ 3.2 & 62.9 $\pm$ 4.3 & 65.9 $\pm$ 3.2 & 400.2 $\pm$ 208 \\
        \midrule
        GRU(2,100) & \textbf{100 $\pm$ 0} & \textbf{99.8 $\pm$ 0.3} & --- & --- & 100 $\pm$ 0 & 69 $\pm$ 4 & --- & --- \\
        DFA extracted & \textbf{99.8 $\pm$ 0.9} & \textbf{99.5 $\pm$ 2.4} & \textbf{99.5 $\pm$ 2.2} &  \textbf{168.1 $\pm$ 261.6} & 76.6 $\pm$ 3.8 & 66.8 $\pm$ 4.3 & 67.8 $\pm$ 4.3 & 389.3 $\pm$ 209 \\
                \midrule
        
        \midrule
        deepDFA(100) & \textbf{100} $\pm$ \textbf{0} & \textbf{99.9} $\pm$ \textbf{0.1} & --- & --- & \textbf{100} $\pm$ \textbf{0} & \textbf{99.9} $\pm$ \textbf{0.2} & --- & --- \\
DFA extracted & \textbf{100} $\pm$ \textbf{0} & \textbf{100} $\pm$ \textbf{0} & \textbf{100} $\pm$ \textbf{0} & \textbf{14.0} $\pm$ \textbf{1.8} & \textbf{99.9} $\pm$ \textbf{0} & \textbf{99.9} $\pm$ \textbf{0.29} & \textbf{99.9} $\pm$ \textbf{0.2} & \textbf{19.2} $\pm$ \textbf{3.0} \\

        \midrule
        \midrule

DFAGen(50) & \textbf{o.o.t.} & ---  & --- & --- & ---  & ---  & --- & --- \\  

        \midrule
DFAGen(100) & \textbf{o.o.t.} & ---  & --- & --- & ---  & ---  & --- & --- \\         
        \bottomrule
    \end{tabular}
    
    \bigskip
        \centering
\setlength{\tabcolsep}{4pt}
    \begin{tabular}{cccccccccc}
        \toprule
        \multirow{2}{*}{Model} & \multicolumn{4}{c}{Random DFA $|Q| = 30$, $|P| = 2$} & \multicolumn{4}{c}{Random DFA $|Q| = 30$, $|P| = 3$} \\
        \cmidrule(lr){2-5} \cmidrule(lr){6-9}
        & Train & Dev & Test & $|\hat{Q}|$ & Train & Dev & Test & $|\hat{Q}|$ \\
        \midrule
        LSTM(1,200) & 99.8 $\pm$ 1.4 & 99.8 $\pm$ 1.2 & --- & --- & 100 $\pm$ 0 & 64.7 $\pm$ 10.4  & --- & --- \\
        DFA extracted & 99.6 $\pm$ 2.6 & 99.4 $\pm$ 3.6 & 99.2 $\pm$ 5.2 & 200.3 $\pm$ 287.9 & 61.2 $\pm$ 7.1 & 59.7 $\pm$ 7.9 & 58.3 $\pm$ 7.4 & 411.7 $\pm$ 192.7 \\
        \midrule
        LSTM(2,200) & 99.9 $\pm$ 0.2 & 98.4 $\pm$ 6.9 & --- & --- & \textbf{100 $\pm$ 0} & \textbf{67.9 $\pm$ 13.1} & --- & --- \\
        DFA extracted & 98.7 $\pm$ 6.2 & 98.3 $\pm$ 7.9 & 98.4 $\pm$ 7.6 & 131.9 $\pm$ 171.9 &  \textbf{64.8 $\pm$ 12.7} & \textbf{61.6 $\pm$ 13.6} & \textbf{61.6 $\pm$ 13.2} & \textbf{350.9 $\pm$ 161.5} \\
        \midrule
        GRU(1,200) & \textbf{99.9 $\pm$ 0.2} & \textbf{99.9 $\pm$ 0.4} & --- & --- & 99.7 $\pm$ 2 & 66.4 $\pm$ 11.9 & --- & --- \\
        DFA extracted & \textbf{99.8 $\pm$ 0.5} & \textbf{99.8 $\pm$ 0.7} & \textbf{99.7 $\pm$ 1} & \textbf{183.1 $\pm$ 266.4} & 62.8 $\pm$ 9.7 & 61.2 $\pm$ 10.6 & 59.6 $\pm$ 10.1 & 420.7 $\pm$ 172.2 \\
        \midrule
        GRU(2,200) & 99.1 $\pm$ 4.7 & 97.9 $\pm$ 7.8 & --- & --- & 100 $\pm$ 0 & 67.6 $\pm$ 12 & --- & --- \\
        DFA extracted & 98 $\pm$ 7.6 & 97.8 $\pm$ 8.4 & 97.6 $\pm$ 9 & 178.5 $\pm$ 244.6 & 63.9 $\pm$ 11.3 & 61.2 $\pm$ 12.3 &  60.1 $\pm$ 11.8 & 383.1 $\pm$ 173.8 \\
                \midrule
        
        \midrule
        deepDFA(200) & \textbf{99.9} $\pm$ \textbf{0.4} & \textbf{99.8} $\pm$ \textbf{0.8} & --- & --- & \textbf{100.0} $\pm$ \textbf{0.0} & \textbf{100.0} $\pm$ \textbf{0.0} & --- & --- \\
DFA extracted & \textbf{99.8} $\pm$ \textbf{0.8} & \textbf{99.8} $\pm$ \textbf{1.1} & \textbf{99.8} $\pm$ \textbf{1.1} & \textbf{22.6} $\pm$ \textbf{1.4} & \textbf{100.0} $\pm$ \textbf{0.0} & \textbf{100.0} $\pm$ \textbf{0.1} & \textbf{100.0} $\pm$ \textbf{0.1} & \textbf{28.1} $\pm$ \textbf{1.2} \\

        \midrule

        \midrule
DFAGen(50) & \textbf{o.o.t.} & ---  & --- & --- & ---  & ---  & --- & --- \\  

        \midrule
DFAGen(100) & \textbf{o.o.t.} & ---  & --- & --- & ---  & ---  & --- & --- \\          
        \bottomrule
    \end{tabular}
    
\end{table*}

\begin{table*}[t!]
    \centering
    \caption{\textbf{L* extraction},\textbf{DeepDFA} and \textbf{DFA-generator} on the \textbf{noisy dataset} constructed for the \textbf{random DFAs} experiment, with different RNN architectures. We highlight in bold the model obtaining the best accuracy on the dev set for each approach.}
    \label{tbl:random_DFA_noisy}
    \centering
\setlength{\tabcolsep}{4pt}
    \begin{tabular}{cccccccccc}
        \toprule
        \multirow{2}{*}{Model} & \multicolumn{4}{c}{Random DFA $|Q| = 10$, $|P| = 2$} & \multicolumn{4}{c}{Random DFA $|Q| = 10$, $|P| = 3$} \\
        \cmidrule(lr){2-5} \cmidrule(lr){6-9}
        & Train & Dev & Test & $|\hat{Q}|$ & Train & Dev & Test & $|\hat{Q}|$ \\
\midrule
LSTM(1,100) & 99.4 $\pm$ 0.2 & 99.6 $\pm$ 0.5 & --- & --- & 99.0 $\pm$ 2.4 & 86.0 $\pm$ 7.4 & --- & --- \\
DFA extracted & 98.3 $\pm$ 5.0 & 98.1 $\pm$ 5.1 & 98.4 $\pm$ 5.0 & 145.9 $\pm$ 239.1 & 94.2 $\pm$ 5.7 & 90.5 $\pm$ 8.2 & 89.3 $\pm$ 10.5 & 286.8 $\pm$ 219.7 \\
\midrule
LSTM(2,100) & \textbf{99.4 $\pm$ 0.4} & \textbf{99.6 $\pm$ 0.5} & --- & --- & \textbf{98.2 $\pm$ 5.5} & \textbf{98.1 $\pm$ 5.6} & --- & --- \\
DFA extracted & \textbf{98.2 $\pm$ 5.5} &\textbf{ 98.1 $\pm$ 5.6} & \textbf{98.4 $\pm$ 5.0} & \textbf{131.5 $\pm$ 225.4} & \textbf{95.8 $\pm$ 4.8} &\textbf{ 92.0 $\pm$ 7.6} & \textbf{92.0 $\pm$ 8.4} & \textbf{290.5 $\pm$ 193.2} \\
\midrule
GRU(1,100) & 99.0 $\pm$ 1.0 & 99.5 $\pm$ 0.7 & --- & --- & 99.3 $\pm$ 1.9 & 84.6 $\pm$ 9.8 & --- & --- \\
DFA extracted & 99.5 $\pm$ 0.7 & 99.5 $\pm$ 0.9 & 99.6 $\pm$ 0.6 & 212.1 $\pm$ 288.4 & 93.0 $\pm$ 6.4 & 87.1 $\pm$ 10.3 & 86.2 $\pm$ 10.9 & 334.2 $\pm$ 163.9 \\
\midrule
GRU(2,100) & 99.1 $\pm$ 1.8 & 99.4 $\pm$ 2.1 & --- & --- & 99.1 $\pm$ 2.4 & 86.5 $\pm$ 12.0 & --- & --- \\
DFA extracted & 97.3 $\pm$ 8.4 & 97.1 $\pm$ 8.5 & 97.5 $\pm$ 8.1 & 138.2 $\pm$ 247.9 & 92.9 $\pm$ 8.5 & 86.8 $\pm$ 13.3 & 86.9 $\pm$ 12.7 & 342.4 $\pm$ 194.5 \\

        \midrule
        
        \midrule
deepDFA(100) & 98.7 $\pm$ 1.1 & 99.5 $\pm$ 1.7 & --- & --- & \textbf{99.0 $\pm$ 0.1} & \textbf{98.3 $\pm$ 1.0} & --- & --- \\
DFA extracted & 93.3 $\pm$ 9.8 & 93.5 $\pm$ 10.7 & 93.6 $\pm$ 10.7 & 11.5 $\pm$ 5.2 & \textbf{96.4 $\pm$ 2.1} & \textbf{95.2 $\pm$ 3.7} & \textbf{95.1 $\pm$ 3.6} & \textbf{37.2 $\pm$ 11.1} \\
        \midrule

        \midrule
DFAGen(50) & \textbf{89.3} \textbf{$\pm$ 20.2} & \textbf{88.6 $\pm$ 20.1} & --- & --- & \textbf{o.o.t.} & --- & --- & --- \\
DFA extracted & \textbf{87} $\pm$ \textbf{18.3} & \textbf{86.6} $\pm$ \textbf{18.3} & \textbf{84} $\pm$ 1\textbf{4.5} & \textbf{15} $\pm$ \textbf{18} & --- & --- & --- & --- \\
        \midrule
DFAGen(100) & \textbf{o.o.t.} & ---  & --- & --- & ---  & ---  & --- & --- \\

        \bottomrule
    \end{tabular}
    \bigskip

        \centering
\setlength{\tabcolsep}{4pt}
    \begin{tabular}{cccccccccc}
        \toprule
        \multirow{2}{*}{Model} & \multicolumn{4}{c}{Random DFA $|Q| = 20$, $|P| = 2$} & \multicolumn{4}{c}{Random DFA $|Q| = 20$, $|P| = 3$} \\
        \cmidrule(lr){2-5} \cmidrule(lr){6-9}
        & Train & Dev & Test & $|\hat{Q}|$ & Train & Dev & Test & $|\hat{Q}|$ \\
\midrule
LSTM(1,100) & 99.2 $\pm$ 1.6 & 97.6 $\pm$ 4.1 & --- & --- & 99.2 $\pm$ 2.5 & 64.5 $\pm$ 3.3 & --- & --- \\
DFA extracted  & 98.3 $\pm$ 6.2 & 97.3 $\pm$ 7.7 & 97.1 $\pm$ 7.7 & 296.0 $\pm$ 244.9 & 71.8 $\pm$ 3.9 & 62.8 $\pm$ 3.5 & 64.5 $\pm$ 4.2 & 389.1 $\pm$ 185.4 \\
\midrule
LSTM(2,100) & 99.3 $\pm$ 1.3 & 98.5 $\pm$ 1.9 & --- & --- & 99.7 $\pm$ 1.2 & 68.1 $\pm$ 3.8 & --- & --- \\
DFA extracted  & 94.4 $\pm$ 14.3 & 93.6 $\pm$ 15.4 & 93.6 $\pm$ 15.3 & 247.2 $\pm$ 263.8 & 74.4 $\pm$ 3.7 & 64.2 $\pm$ 4.6 & 67.4 $\pm$ 5.1 & 389.2 $\pm$ 216.0 \\
\midrule
GRU(1,100) & 99.1 $\pm$ 1.7 & 97.1 $\pm$ 5.1 & --- & --- & 98.5 $\pm$ 3.5 & 64.5 $\pm$ 4.0 & --- & --- \\
DFA extracted  & 98.9 $\pm$ 2.9 & 97.6 $\pm$ 5.7 & 97.6 $\pm$ 5.7 & 334.7 $\pm$ 273.9 & 71.0 $\pm$ 2.8 & 62.5 $\pm$ 3.6 & 64.5 $\pm$ 3.0 & 386.6 $\pm$ 193.1 \\
\midrule
GRU(2,100) & \textbf{99.6 $\pm$ 0.2} & \textbf{98.9 $\pm$ 1.1} & --- & --- & \textbf{99.3 $\pm$ 2.2} & \textbf{68.4 $\pm$ 3.9} & --- & --- \\
DFA extracted  & \textbf{99.5 $\pm$ 1.0} & \textbf{98.8 $\pm$ 2.2} & \textbf{98.9 $\pm$ 1.9} & \textbf{359.0 $\pm$ 251.2} & \textbf{73.5 $\pm$ 4.3} & \textbf{64.2 $\pm$ 4.4} & \textbf{66.2 $\pm$ 4.5} & \textbf{420.3 $\pm$ 220.3} \\

                \midrule
        
        \midrule
deepDFA(100) & \textbf{99.0 $\pm$ 0} & \textbf{99.8 $\pm$ 0.2} & --- & --- & \textbf{99.0 $\pm$ 0.1} & \textbf{96.6 $\pm$ 1.2} & --- & --- \\
DFA extracted & \textbf{98.8 $\pm$ 0.6} & \textbf{99.6 $\pm$ 1.6} & \textbf{99.6 $\pm$ 1.7} & \textbf{14.7 $\pm$ 3.2} & \textbf{97.7 $\pm$ 1.1} & \textbf{96.4 $\pm$ 3.4} & \textbf{96.0 $\pm$ 4.1} & \textbf{34.9 $\pm$ 10.1} \\

        \midrule

        \midrule
DFAGen(50) & \textbf{o.o.t.} & ---  & --- & --- & ---  & ---  & --- & --- \\  

        \midrule
DFAGen(100) & \textbf{o.o.t.} & ---  & --- & --- & ---  & ---  & --- & --- \\    
        \bottomrule
    \end{tabular}
    
    \bigskip
        \centering
\setlength{\tabcolsep}{4pt}
    \begin{tabular}{cccccccccc}
        \toprule
        \multirow{2}{*}{Model} & \multicolumn{4}{c}{Random DFA $|Q| = 30$, $|P| = 2$} & \multicolumn{4}{c}{Random DFA $|Q| = 30$, $|P| = 3$} \\
        \cmidrule(lr){2-5} \cmidrule(lr){6-9}
        & Train & Dev & Test & $|\hat{Q}|$ & Train & Dev & Test & $|\hat{Q}|$ \\
\midrule
LSTM(1,100) &\textbf{ 98.9 $\pm$ 3.8} & \textbf{98.0 $\pm$ 6.8} & --- & --- & 99.2 $\pm$ 1.9 & 64.0 $\pm$ 9.4 & --- & --- \\
DFA extracted & \textbf{98.3 $\pm$ 7.1} & \textbf{98.4 $\pm$ 6.8} & \textbf{98.2 $\pm$ 7.4} &\textbf{ 238.9 $\pm$ 252.2} & 62.1 $\pm$ 8.6 & 60.2 $\pm$ 9.3 & 58.4 $\pm$ 8.7 & 397.4 $\pm$ 166.0 \\
\midrule
LSTM(2,100) & 98.6 $\pm$ 3.0 & 93.7 $\pm$ 11.9 & --- & --- & \textbf{99.8 $\pm$ 0.9} & \textbf{67.1 $\pm$ 11.6} & --- & --- \\
DFA extracted & 94.7 $\pm$ 11.0 & 93.9 $\pm$ 12.0 & 93.4 $\pm$ 13.1 & 211.2 $\pm$ 189.9 & \textbf{63.6 $\pm$ 10.9} & \textbf{61.6 $\pm$ 11.5} & \textbf{60.4 $\pm$ 11.3} & \textbf{338.7 $\pm$ 173.4} \\
\midrule
GRU(1,100) & 98.3 $\pm$ 5.2 & 95.9 $\pm$ 10.8 & --- & --- & 98.6 $\pm$ 2.8 & 65.1 $\pm$ 10.4 & --- & --- \\
DFA extracted & 96.4 $\pm$ 9.8 & 95.9 $\pm$ 10.9 & 95.7 $\pm$ 11.2 & 248.4 $\pm$ 238.3 & 61.3 $\pm$ 8.1 & 60.2 $\pm$ 9.5 & 59.2 $\pm$ 8.1 & 391.0 $\pm$ 182.3 \\
\midrule
GRU(2,100) & 99.1 $\pm$ 1.3 & 94.1 $\pm$ 11.6 & --- & --- & 99.8 $\pm$ 0.8 & 66.0 $\pm$ 11.4 & --- & --- \\
DFA extracted & 95.1 $\pm$ 10.6 & 94.3 $\pm$ 12.2 & 94.0 $\pm$ 12.8 & 253.1 $\pm$ 235.4 & 62.9 $\pm$ 9.4 & 60.1 $\pm$ 10.9 & 59.5 $\pm$ 9.6 & 378.5 $\pm$ 167.6 \\

                \midrule
        
        \midrule
deepDFA(200) & \textbf{98.9 $\pm$ 0.4} & \textbf{99.8 $\pm$ 0.7} & --- & --- & \textbf{98.9 $\pm$ 0.1} & \textbf{99.2 $\pm$ 1.2} & --- & --- \\
DFA extracted & \textbf{98.8 $\pm$ 0.9} & \textbf{99.7 $\pm$ 1.0} & \textbf{99.6 $\pm$ 1.5} & \textbf{22.6 $\pm$ 1.7} & \textbf{98.2 $\pm$ 1.2} & \textbf{98.7 $\pm$ 2.0} & \textbf{98.7 $\pm$ 2.0} & \textbf{55.1 $\pm$ 34.0} \\

        \midrule

        \midrule
DFAGen(50) & \textbf{o.o.t.} & ---  & --- & --- & ---  & ---  & --- & --- \\  

        \midrule
DFAGen(100) & \textbf{o.o.t.} & ---  & --- & --- & ---  & ---  & --- & --- \\    
        \bottomrule
    \end{tabular}
    
\end{table*}

\subsection{Complete DeepDFA and L* Extraction Results}
This section presents comprehensive results for the DeepDFA and L* extraction methods. We report the train and dev accuracies for all RNNs and DeepDFA models used in this experiment. Additionally, we showcase the train, deve, and test accuracies of the DFA extracted from the continuous model, alongside the number of states ($|\hat{Q}|$) of the extracted DFA. For RNNs, the DFA is extracted using the L* extraction algorithm. In the case of DeepDFA models, the crisp DFA is extracted by evaluating the parameters with minimum temperature activation. For each language and neural model, we conduct $N$ runs with varying random seeds, reporting the mean $\pm$ standard deviation. Specifically, we set $N=5$ for the Tomita Languages and $N=10$ for the random DFAs. The results are presented in Tables \ref{tbl:L_star_tomita} to \ref{tbl:random_DFA_noisy}.

In these tables, we highlight in bold the performance metrics that correspond to models achieving \textit{perfect results}. A model is deemed to achieve perfect performance when the continuous model attains 100\% accuracy on both the trai and dev sets, the extracted DFA achieves 100\% accuracy on the train, dev, and test sets, and the predicted DFA size matches the target size.

If none of the RNNs or DeepDFA models can achieve perfect performance for a given language, we emphasize in bold the model with the best dev accuracy. If two models share the same dev accuracy, we select the one with the number of states closest to the target. These highlighted models are the ones showcased in the main paper results. In particular, in that results we selected the model with fewer parameters in the case of multiple models achieving perfect performance.

In the tables, we denote RNN architectures as $<$RNN\_type$>$(l,h), where $l$ is the number of layers, $h$ is the hidden state size, and $<$RNN\_type$>$ represents the type of RNN tested. For DeepDFA architectures, we use DeepDFA(h), where $h$ is the maximum number of states ($|\hat{Q}_{max}|$).

\subsubsection{L* Extraction on Tomita Datasets (Perfectly Labeled)}
Table \ref{tbl:L_star_tomita} presents the results for the experiments on Tomita Languages without label noise.

\paragraph{T1 and T2:} For the simplest Tomita languages (T1 and T2), we observe that 3 out of 6 models achieve perfect performance for T1, and 2 out of 6 for T2. Interestingly, the results demonstrate that adding weights doesn't consistently improve performance. For instance, T1 shows better results with a one-layer GRU with a hidden state size of 30 compared to a larger one-layer GRU with a hidden state size of 100. Regarding T2, the best outcomes are obtained with shallow one-layer RNN architectures, while two-layer architectures struggle to achieve perfection due to the predicted DFA having too few states, causing a slight accuracy reduction on the training set for both LSTM and GRU.

\paragraph{T2 and T3:} Despite T2 and T3 being of medium difficulty, they are solved more easily compared to T1 and T2. In these cases, all architectures except one (GRU(1, 100)) achieve perfect performance. However, the single exception yields a predicted DFA with an exceptionally larger number of states (285 instead of 5) for T2, indicating a significant deviation from the target DFA.

\paragraph{T5 and T6:} T5 and T6 prove to be the most challenging languages in the benchmark. For both languages, none of the RNNs achieve perfect performance when considering both the RNN and the extracted DFA. Nonetheless, for T5, perfect results are attainable with the DFA and nearly perfect accuracy with the RNN when using GRU(2, 100). However, for T6, no tested models achieve 100\% accuracy with either the RNN or the DFA. In this case, we select the best model based on the highest dev RNN accuracy: LSTM(1, 100). It's noteworthy that the extracted DFA for T6 consistently contains many more states than the target (minimum of 42 and hundreds of states in other cases).

\paragraph{T7:} Despite theoretically being the most challenging language in the benchmark, all RNNs achieve perfect performance on T7, leading us to omit the results for space considerations.

\subsubsection{L* Extraction on Noisy Tomita Datasets}
Tables \ref{tbl:L_star_Tomita_noisy} and \ref{tbl:T7_L*_noisy} present results for the noisy versions of the Tomita datasets, where 1\% of training labels were flipped.

\paragraph{T1 and T2:} For T1, 4 out of 6 models predicts the DFA perfectly, while for T2, 2 out of 6 models achieve perfect performance. It's important to note that we consider models with the same number of states as the target to be perfect, but this information isn't available during testing. If we selected the model with the highest dev accuracy as the best, we would also choose LSTM(1, 30) for T1. However, this model's predicted number of states (124) significantly deviates from the actual number. A similar situation arises for T2, where models with non-target state numbers achieve perfect dev accuracy, making them indistinguishable from truly perfect models.

\paragraph{T3 and T4:} Introducing errors significantly decreases the number of models achieving perfect performance. None of the tested models achieve perfect results for T3, while only 2 out of 6 models succeed for T4, in contrast to the perfect performances obtained without noise.

\paragraph{T5 and T6:} The best models from the noiseless scenario remain the best in the noisy setting as well (GRU(2, 100) for T5 and LSTM(1, 100) for T6).

\paragraph{T7:} Despite all 6 models achieving perfect performance without noise, only 2 models (GRU(1, 100) and GRU(2, 100)) maintain perfect performance.

\subsubsection{DeepDFA on Tomita Datasets (Perfectly Labeled)}
Table \ref{tbl:deepDFA_Tomita} presents DeepDFA results on Tomita Languages using different state sizes (10, 30, and 100).

In the majority of experiments, DeepDFA achieves perfect results, exhibiting 100\% accuracy on all three datasets (training, development, and test) for both the continuous model and the DFA extracted at the minimum temperature activation, and a DFA size ($|\hat{Q}|$) equal to the target size.

Specifically, all tested models achieve perfection for languages T1, T2, T4, T5, and T6. For T3, one model out of three achieves perfect performance, and for T7, 2 out of 3 models achieve perfection. Importantly, when the target DFA cannot be perfectly inferred, the predicted DFA's number of states deviates from the target by less than one state, in contrast to L* extraction, which tends to significantly overestimate the number of states.

\subsubsection{DeepDFA on Noisy Tomita Datasets}
Table \ref{tbl:deepdfa_Tomita_noisy} showcases results for DeepDFA on the Tomita Languages with 1\% label noise. DeepDFA maintains perfect results regardless of the maximum number of states chosen for Tomita 4 and 6. For Tomita 1, one model out of three predicts the DFA perfectly. For other languages (Tomita 2, 3, 5, and 7), the majority of models (2 out of 3 tested) achieve perfect performance. Notably, even when comparing the worst performance of DeepDFA on these datasets, it is much closer to the optimal outcome than the worst performance of L* extraction, especially regarding state estimation.

\subsubsection{L* Extraction and DeepDFA on Random DFA Datasets (Perfectly Labeled)}
Table \ref{tbl:random_dfa} presents results for L* extraction and DeepDFA on random DFA datasets without noise.

Regarding \textbf{L* extraction}, several observations are noteworthy:
\begin{itemize}
    \item 
Overfitting is prominent in some cases, particularly for target DFAs with sizes $>$ 10 and symbol counts $>$ 2. Although the RNN's training accuracy approaches 100\%, the development accuracy is around 60-70\%. Reducing RNN dimensions in these cases to mitigate overfitting results in similar performance.
    \item 
Significant performance loss due to abstraction is evident. This phenomenon also appears in Tomita language experiments (where the DFA's accuracy on the training set is slightly lower than that of the RNN). However, the difference is more pronounced here, especially for target DFAs with $|Q| > 10$ and $|P| > 2$.
    \item 
Predicted DFA sizes are consistently much larger than the target, even for the best-performing tested model.
\end{itemize}

The first two observations are likely connected, since they occur on the same DFAs.

Regarding \textbf{DeepDFA} (last row of the table), even with only one model architecture tested, DeepDFA outperforms all 4 tested RNNs across all random DFAs. Notably, there is no overfitting, and discretization-related accuracy loss is nearly negligible in most cases, with a loss of 2 accuracy points in the worst case scenario.

\subsubsection{L* Extraction and DeepDFA on Noisy Random DFA Datasets}
Table \ref{tbl:random_dfa} presents results for L* extraction and DeepDFA on random DFA datasets with 1\% label noise.

Across most configurations of random DFAs, DeepDFA achieves better performance than the best RNN. In one instance ($|Q|=10$, $|P|=2$), L* extraction achieves higher accuracy but overestimates the number of states by over ten times compared to the target. In contrast, DeepDFA's accuracy is slightly lower but its predicted number of states (11.5) closely aligns with the target (10). Both methods achieve performance similar to the noiseless scenario. Notably, DeepDFA overestimates the number of states to a greater extent when $|P| = 3$ compared to the noiseless scenario, but this overestimation is still much lower than that of L* extraction.

\end{document}